\documentclass{article}

\usepackage[preprint]{neurips_2026}

\usepackage[utf8]{inputenc} 
\usepackage[T1]{fontenc}    
\usepackage{hyperref}       
\usepackage{url}            
\usepackage{booktabs}       
\usepackage{amsfonts}       
\usepackage{nicefrac}       
\usepackage{microtype}      
\usepackage{xcolor}         

\usepackage{amsmath}
\usepackage{graphicx}
\usepackage{multirow}
\usepackage{array}
\usepackage{wrapfig}

\title{RotMoLE: Enhancing Mixture of Low-Rank Experts through Rotational Gating Mechanism}

\author{
    \textbf{Mengyang Sun}$^1$ \thanks{ The work was done while the author interned at Zhipu.AI.} \quad 
    \textbf{Maochuan Dou}$^2$ \quad 
    \textbf{Tao Feng}$^1$ \quad 
    \textbf{Dan Zhang}$^3$ \\[0.6em]
    \textbf{Yihao Wang}$^2$ \quad  
    \textbf{Junpeng Liu}$^4$ \quad 
    \textbf{Yifan Zhu}$^5$ \quad 
    \textbf{Jie Tang}$^1$ \\[0.6em]
    \small $^1$Tsinghua University \quad 
    \small $^2$Beijing Information Science and Technology University \and 
    \small $^3$National University of Singapore \quad
    \small $^4$Hong Kong University of Science and Technology (Guangzhou) \and
    \small $^5$Beijing University of Posts and Telecommunications \\
    \small \texttt{sunmy19@mails.tsinghua.edu.cn} \quad
    \small \texttt{\{2023011446, 2022011219\}@bistu.edu.cn} \and
    \small \texttt{fengtao.hi@gmail.com} \quad
    \small \texttt{zhangdan25@nus.edu.sg} \quad
    \small \texttt{junpengliu@hkust-gz.edu.cn} \\
    \small \texttt{yifan\_zhu@bupt.edu.cn} \quad
    \small \texttt{jietang@tsinghua.edu.cn}
}

\begin{document}

\maketitle

\begin{abstract}
  While Large Language Models (LLMs) are commonly fine-tuned to handle domain-specific tasks before being applied to vertical applications, adapting them to complex scenarios with diverse specialized knowledge remains challenging. Meanwhile, Mixture-of-Experts (MoE) architecture has risen as a crucial paradigm for training LLMs, and some recent works have also incorporated MoE into Parameter-Efficient Fine-Tuning (PEFT) to propose the Mixture of Low-rank Experts (MoE-LoRA), to enhance the power of low-rank adapters for learning complicated knowledge. However, conventional gating mechanisms in MoE typically apply only a scalar reweighing to selected experts, thereby limiting their underlying capacity of representation and generalization. Motivated and enabled by the low-rank structures in MoE-LoRA, we propose RotMoLE, a specialized MoE framework for low-rank experts featuring an additional rotation gate. Beyond simple scaling, RotMoLE implements a rotation mechanism for each selected expert, enabling superior expert exploitation and specialization for learning diverse data, especially when expert candidates are limited. Empirical results on complex multi-task and multilingual training scenarios validate our effectiveness.
\end{abstract}


\section{Introduction}
\label{sec:introduction}

Through Mixture-of-Experts (MoE)~\citep{jacobs1991adaptive} architecture and Parameter-Efficient Fine-Tuning (PEFT)~\citep{zhang2025parameter} technique, large language models (LLMs) exhibit their capabilities of accommodating massive general knowledge and rapidly adapting to specific vertical domains. As noted by \citet{cai2025survey} and \citet{zhang2025mixture}, MoE serves as a sparse framework for efficiently scaling LLMs. Through top-$k$ selection and weighted aggregation of experts, MoE exhibits its significant capacity and flexibility across diverse fields, especially for complex and multi-task scenarios. Specifically, top-$k$ expert selection in MoE enables sparse activation, which thereby not only boosts the computational efficiency~\citep{shazeer2017outrageously}, but also facilitates the decoupling and isolation of specialized domain knowledge~\citep{ren2023pangu}. Additionally, the weighted expert aggregation acts as a magnitude scaling process applied to the outputs of selected experts based on their respective gate values, thereby further expanding the representational capacity of MoE~\citep{nie2021dense}. The whole paradigm is formulated as $y = \sum\nolimits_{i\,\in\,TopK(x)} g_iE_i(x)$, where $g_i$s denote the gate values and we typically normalize $\sum_{i\,\in\,TopK(x)} g_i = 1$ to ensure numerical stability. 

Furthermore, as the Low-rank Adaptation (LoRA)~\citep{hu2022lora} gains increasing traction in the field of PEFT, recent researchers have also incorporated MoE into LoRA and proposed Mixture of Low-rank Experts (MoE-LoRA)~\citep{dou2024loramoe,li2024mixlora}. Specifically, conventional LoRA introduces a trainable pair of low-rank matrices $A_{r \times d}$ and $B_{d \times r}$ ($r \ll d$) to a fixed pretrained module $W_{d \times d}$, given by $W=W_0+BA$; while MoE-LoRA similarly introduces a series of LoRA modules $B_i$ and $A_i$, as well as a routing module $g=G(x)$ to W, given by $W=W_0+\sum_{i\,\in\,TopK(x)}g_iB_iA_i$. The incorporation of multiple low-rank modules in MoE-LoRA demonstrably improved the performance of PEFT in various scenarios, such as multi-task learning~\citep{liu2023moelora} and continual learning~\citep{gedynamic}, etc. However, there still remain challenges in modeling diverse specialized knowledge in complicated downstream scenarios using such parameter-efficient methods like MoE-LoRA. 

\begin{figure}[t]
\centering
\includegraphics[width=\linewidth]{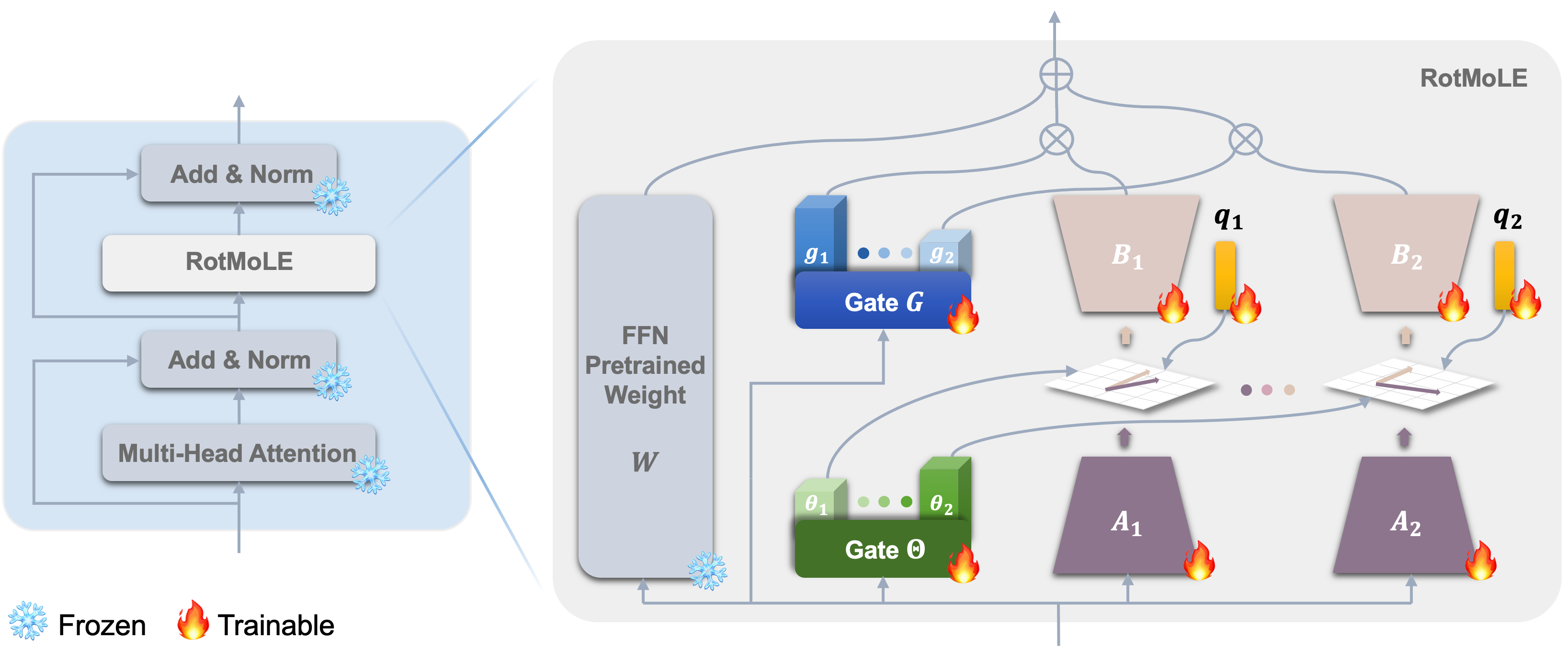}
\caption{The structure of our Rotatable Mixture of Low-rank Experts (i.e., RotMoLE): To model a rotation operation for each expert beyond the simple gate-value-based scaling, we implement a rotation gate $\theta=\Theta(x)$ as well as a set of learnable vectors $\boldsymbol{q}$, for generating rotation angles and rotation planes for each low-rank expert. Specifically, for the LoRA expert's rank equal to $r$, we conduct a 2D rotation for the $r$-dimensional embeddings in a 2D sub-plane, given by $W=W_0+\sum_{i\,\in\,TopK(x)}g_iB_iR_iA_i$, where $R_i$ denotes the $r$-dimensional rotation matrix constructed by $\theta_i$ and $\boldsymbol{q}_i$.}
\vspace{-10pt}
\label{figure0}
\end{figure}

Most MoE frameworks employ a simple scaling gate to select and weigh experts, which effectively modulates the magnitude of experts' respective contributions. However, it fundamentally lacks the capacity to perform complex spatial transformations on those selected experts. Consequently, it is insufficient for modeling intricate exploitation strategies of candidate experts required by diverse and complicated scenarios, such as multilingual or multi-task learning, especially when candidate experts are limited. Specifically, scalar gate values are only able to scale the norms of expert outputs along their fixed spatial directions. This limitation restricts the solution space formed by the weighted aggregation of selected experts. As a result, those routers are only limited to determining \emph{how much} an expert contributes, while overlooking \emph{how diverse} it can contribute. 

A natural solution is to introduce an enhanced gate module capable of providing each selected expert with a more powerful transformation function beyond scaling. However, this brings a substantial demand for additional trainable parameters. For example, an unconstrained linear transformation of a $d$-dimensional vector requires a $d \times d$ matrix. Therefore, with $n$ $d$-dimensional experts, this would rely on $d^3n$ trainable gating parameters, which are disproportionately large for a routing module, and may even be far larger than expert modules themselves. Dynamically constructing these matrices from fewer coefficients during inference may alleviate the storage burden, but it still leads to a temporary $d \times d$ memory usage and a computational cost of $d$-dimensional matrix construction and multiplication. Given that the hidden dimension $d$ in current mainstream LLMs typically exceeds $4k$~\citep{zhao2023survey}, such an approach is effectively prohibitive compared to its benefits. Notably, the low-rank structure of LoRA offers an opportunity to potentially implement complex expert transformations in MoE-LoRA that go beyond simple scaling: Since $r \ll d$, an $r$-dimensional linear transformation function requires only an $r \times r$ matrix if we apply it after module A, such as $W=W_0+\sum_{i\,\in\,TopK(x)}B_iG_iA_i$, where $G_i$ is an $r \times r$ matrix generated by a gate module with only $dnr^2$ trainable parameters. However, compared to the low-rank expert modules themselves which totally consist of $2ndr$ parameters, it is still a disproportionately large gate.

Motivated by the observation that any transformation between two specific non-zero vectors can be formulated as a 2D transformation within the sub-plane spanned by the two vectors, and thus can be further decomposed into a pure scaling operation and a pure rotation within the plane, we propose Rotatable Mixture of Low-rank Experts (i.e., RotMoLE), a specialized MoE framework for low-rank experts featuring an additional rotation gate. Specifically, RotMoLE assigns not only a scaling value $g_i$, but also a rotation coefficient $\theta_i$ to each candidate expert. For the case where the expert rank $r$ equals 2, a 2D rotation matrix can be naturally derived as:

$$R_i = \left(
\begin{array}{rr}
    \cos{\theta_i} & -\sin{\theta_i} \\
    \sin{\theta_i} & \cos{\theta_i}
\end{array}
\right);$$

While for $r$ larger than 2, the rotation operation is much more complex to formulate since it requires more variables beyond a single $\theta_i$ to specify the rotation sub-plane $\mathcal{P}_i$ for expert $i$. Therefore, we further employ a small set of $rn$ trainable parameters to determine a temporary rotation plane $\mathcal{P}_i$ for each expert and thereby construct the $r \times r$ rotation matrix $R_i$ based on $\mathcal{P}_i$ and $\theta_i$. As a result, our proposed RotMoLE can be expressed as $W=W_0+\sum_{i\,\in\,TopK(x)}g_iB_iR_iA_i$, while the total number of extra parameters we introduce is only $dn$ for LoRA ranks equal to 2, and $(d+r)n$ for LoRA ranks larger than 2, respectively. The main structure of RotMoLE is exhibited by Figure \ref{figure0}.

The main concepts of this paper are related to the fields of MoE in LLMs, PEFT, LoRA, and MoE-LoRA, etc. Please refer to Appendix \ref{apdx_related_work} for related works of those fields. Our main contributions are summarized as follows:

\begin{itemize}
\item We emphasize the limitation of existing MoE routers, which only perform a norm scaling for each selected expert, lacking the capability to perform complex transformations on experts for diverse scenarios, especially when experts are limited.
\item We indicate the potential of implementing more complex transformations for routing low-rank experts in MoE-LoRA, and propose RotMoLE as a specialized MoE-LoRA framework featuring an additional rotation gate.
\item We conduct a series of multi-task and multilingual experiments under various circumstances, illustrating the effectiveness and generalization of our RotMoLE over conventional MoE-LoRA baselines.  
\end{itemize}




\section{Methodology}
\label{method_sec}

\begin{wrapfigure}{r}{0.5\textwidth}
\centering
\includegraphics[width=\linewidth]{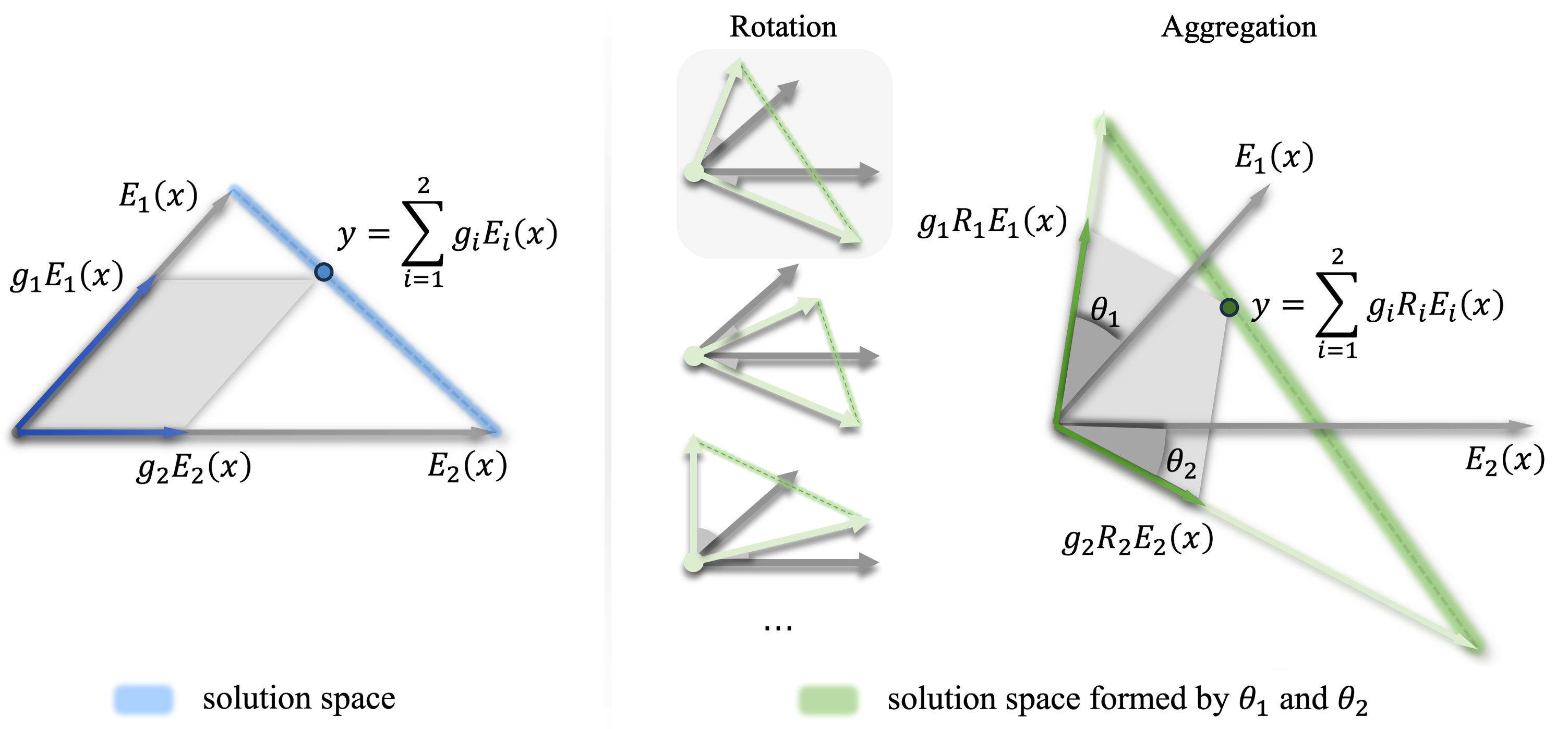}
\caption{Illustrations of the MoE solution space through a case study in a $2$-dimensional space with two candidate experts.} 
\label{figure1}
\end{wrapfigure}

Our RotMoLE serves as a specialized framework of MoE-LoRA, relying on decomposing an unconstrained transformation of candidate experts into a pure scaling operation and a pure rotation within a $2$-dimensional plane space. Specifically, we treat the conventional gate modules in MoE as the scaling operators, so the weighted aggregation based on their generated gate values can be regarded as a normalization to preserve the scale consistency of hidden states during layer-up computation; To model rotation, we introduce an extra gate module named rotation gate, which generates rotation angle values $\theta$s for selected experts. A specific low-rank rotation matrix $R_i$ is subsequently derived from the respective angle value $\theta_i$ for each expert $i$, and then multiplying the expert in the middle of two low-rank matrices $A_i$ and $B_i$, given by $W=W_0+\sum_{i\,\in\,TopK(x)}g_iB_iR_iA_i$. We claim that performing customized rotation operations for each expert can further expand the global solution space of MoE. We illustrate this through a simple case study in Figure \ref{figure1}. The left part of Figure \ref{figure1} shows the solution space of a conventional 2-expert MoE structure: Given the outputs of candidate experts $E_1(x)$ and $E_2(x)$, and their respective gate values $g_1$ and $g_2$ ($g_1+g_2=1$), the weighted-sum solution space locates only within the blue line; While the right part of Figure \ref{figure1} demonstrates the case of integrating a rotation gate (which is our proposed method): Given respective rotation angles $\theta_1$ and $\theta_2$, outputs of candidate experts are rotated by $\theta_i$ first, and then weighted summed by their respective gate values. The solution space of weighted-summing thus transfers to the green line. As a result, through dynamic and customized $\theta$s, the solution space of MoE is expanded.

We elaborate on our RotMoLE in the following four aspects: (1) the theoretical foundation of RotMoLE; (2) the rotation gate module generating angle values; (3) the algorithm of conducting low-rank rotation through rotation matrix, typically for expert ranks equal to 2; and (4) the solution we propose for rotation matrix construction for larger expert ranks.

\subsection{Theoretical Foundation}
\label{theory}

Our RotMoLE is based on our observation that any transformation between two specific non-zero vectors, no matter originally linear or non-linear, can be reversely written as a 2D transformation within the sub-plane spanned by the two vectors, and thus can be further decomposed into a pure scaling operation and a pure rotation within the plane. To specify this, we first denote the two random non-zero $d$-dimensional vectors as $\boldsymbol{u}$ and $\boldsymbol{v}$, and two orthogonal basis vectors of the plane $\mathcal{P}$ spanned by $\boldsymbol{u}$ and $\boldsymbol{v}$ as $\boldsymbol{e_1}$ and $\boldsymbol{e_2}$, where $|\boldsymbol{e_1}|=|\boldsymbol{e_2}|=1$ and $\boldsymbol{e_1}\cdot\boldsymbol{e_2}=0$. Then we can rewrite $\boldsymbol{u}$ and $\boldsymbol{v}$ as:

$$\boldsymbol{u} = (\boldsymbol{u}\cdot\boldsymbol{e_1})\boldsymbol{e_1} + (\boldsymbol{u}\cdot\boldsymbol{e_2})\boldsymbol{e_2} = u_1\boldsymbol{e_1} + u_2\boldsymbol{e_2};$$
$$\boldsymbol{v} = (\boldsymbol{v}\cdot\boldsymbol{e_1})\boldsymbol{e_1} + (\boldsymbol{v}\cdot\boldsymbol{e_2})\boldsymbol{e_2} = v_1\boldsymbol{e_1} + v_2\boldsymbol{e_2},$$

and so that we can define $\boldsymbol{u}_\mathcal{P} = [u_1, u_2]^T$ and $\boldsymbol{v}_\mathcal{P} = [v_1, v_2]^T$ as the views of $\boldsymbol{u}$ and $\boldsymbol{v}$ under the condition of plane $\mathcal{P}$. Afterwards, these two vectors can be further expressed by:

$$\boldsymbol{u}_\mathcal{P}= \sqrt{u_1^2+u_2^2} \cdot [\cos{\theta_u}, \sin{\theta_u}]^T; \boldsymbol{v}_\mathcal{P}= \sqrt{v_1^2+v_2^2} \cdot [\cos{\theta_v}, \sin{\theta_v}]^T,$$

where $\theta_u = \arccos{\frac{u_1}{\sqrt{u_1^2+u_2^2}}}$ and $\theta_v = \arccos{\frac{v_1}{\sqrt{v_1^2+v_2^2}}}$. Finally, we can formulate the transformation from $\boldsymbol{u}_\mathcal{P}$ to $\boldsymbol{v}_\mathcal{P}$ as:

\begin{equation}
\begin{aligned}
\boldsymbol{v}_\mathcal{P} = \frac{\sqrt{v_1^2+v_2^2}}{\sqrt{u_1^2+u_2^2}} \cdot Rotate(\boldsymbol{u}_\mathcal{P}, \theta_v - \theta_u) = \frac{\sqrt{v_1^2+v_2^2}}{\sqrt{u_1^2+u_2^2}} \cdot \begin{pmatrix}
 \cos{(\theta_v - \theta_u)} & -\sin{(\theta_v - \theta_u)}\\
 \sin{(\theta_v - \theta_u)} & \cos{(\theta_v - \theta_u)}
\end{pmatrix} \cdot \boldsymbol{u}_\mathcal{P}, \notag \\
\end{aligned}
\end{equation}

indicating that we first rotate the vector $\boldsymbol{u}_\mathcal{P}$ by the angle of $\theta_v - \theta_u$ in plane $\mathcal{P}$, and then scale it by a scalar $\frac{\sqrt{v_1^2+v_2^2}}{\sqrt{u_1^2+u_2^2}}$.

\subsection{Rotation Gate}

Similar as the scaling gate $g = G(x) = Softmax(x \cdot W_g)$, we also implement a linear module for the rotation gate. Here we denote it as $x \cdot W_\theta$. Instead of employing a $Softmax$ function which targets for a sum-to-1 normalization among all $n$ experts, our rotation operation naturally does not require a sum-to-1 normalization since the operation of rotating each expert is respectively independent. However, we notice that the rotation angles exhibit periodicity. This non-monotonic property may significantly affect model training. Therefore, we implement an independent normalization for each angle to restrict it within a single period. The whole rotation gate is expressed as follows:

$$\theta = \Theta(x) = 2 \pi \cdot Sigmoid(x \cdot W_\theta) - \pi,$$

where the $Sigmoid$ function normalizes per-expert value to $[0,1]$, and then it is mapped to $[-\pi, \pi]$. We initialize the trainable parameters $W_\theta$ to $0$ since we regard the default condition as the condition without any rotation. Specifically, $Sigmoid$ function is more sensitive to input variations within the nearby range of $0$, while more stable to larger positive or negative inputs. As a result, $\theta$ is sensitive to trainable parameters $W_\theta$ around $2 \pi \cdot Sigmoid(0) - \pi = 0$, while more stable around $-\pi$ or $\pi$ when $W_\theta$ is relatively large. This will theoretically encourage both early-stage convergence and late-stage stability of our rotation gate, boosting the training performance of RotMoLE.

For sparse expert activation, instead of conducting top-$k$ expert selection based on rotation angle values, we still follow the convention of previous MoE frameworks to perform sparse activation only based on scaling gate values $g_i$s. In other words, $k$ experts with the largest $g_i$ values are activated. This is due to the concept that only scaling gate values $g_i$s indicate the relative importance of each expert, while the rotation gate only contributes to the diversity of expert exploitation and generalization. 

\subsection{2-Dimensional Rotation}

Two essential components for conducting a rotation operation in a $d$-dimensional space are the rotation angle and the rotation plane (i.e., a $2$-dimensional subspace). For the case where expert rank equals $2$, the entire space is $2$-dimensional, so the rotation angle serves as the only coefficient for modeling a rotation operation. A $2$-dimensional rotation can be expressed by a $2$-dimensional matrix $R_i$, and therefore, our proposed RotMoLE can be derived as the follows:

\begin{equation}
\begin{aligned}
W=W_0+\textstyle\sum_{i\,\in\,TopK(x)}g_iB_iR_iA_i=W_0+\textstyle\sum_{i\,\in\,TopK(x)}g_iB_i\begin{pmatrix}
 \cos{\theta_i} & -\sin{\theta_i} \\
 \sin{\theta_i} & \cos{\theta_i}
\end{pmatrix}A_i, \notag
\end{aligned}
\end{equation}

where $g=G(x)$ and $\theta=\Theta(x)$, indicating the scaling and the rotation gate modules, respectively.

\subsection{Rotation for Larger LoRA Ranks}
\label{rotation_for_larger_ranks}

To implement rotation operation for expert rank $r$ larger than 2, we follow the theoretical foundation in Section \ref{theory}, which indicates that the $r$-dimensional rotation operation inside each $r$-rank LoRA expert can be formulated as a $2$-dimensional rotation by constructing a rotation plane $\mathcal{P}_i(x)$ for each expert $i$ and each query $x$, spanned by the input embedding $\boldsymbol{u}$ (i.e., $A_ix$, since $R_i$ is in the middle of $A_i$ and $B_i$) and output embedding $\boldsymbol{v}$. However, as the rotation output $\boldsymbol{v}$ is unknown and is just what we aim to obtain, this rotation plane can not be determined only by $\boldsymbol{u}$. As a result, we also need to model rotation planes as variables, just as the way we model rotation angles. 

Specifically, as in expert $i$ the input embedding $\boldsymbol{u}$ for rotation (i.e., $A_ix$) definitely lies in its rotation plane, we consider one of the plane basis $\boldsymbol{e_1}$ as $\frac{A_ix}{|A_ix|}$, leaving the other $r$-dimensional basis $\boldsymbol{e_2}$ as the only variable. Then, instead of applying another input-related gate to generate a specific $\boldsymbol{e_2}$ for each expert through a linear mapping $x_{d \times 1} \to n{\boldsymbol{e_2}}_{r \times 1}$ which requires $dnr$ extra trainable parameters, we implement it in a lighter way. Specifically, \citet{luo2024moelora} claims that since each expert is designed for specializing in a specific domain of knowledge, input embeddings routed to the same expert are expected to be more similar than those routed to distinct experts, and so are their output embeddings. Based on that claim, the inputs $x$s routed to the same LoRA expert $i$ are similar and can be regarded as a relatively aggregated cluster, and so are the $A_ix$ embeddings outputted by matrix $A_i$. As a result, the dispersion level of this $A_ix$s cluster indicates the diversity within expert $i$, and finally their respective rotation operations can therefore be light-weighted as functions enhancing this diversity to achieve better expert exploitation. To model this diversity enhancement, we constrain and simplify our rotation operation in expert $i$ as a rotation only towards or away from the center of expert $i$'s cluster, which also means that we require the rotation plane $\mathcal{P}_i(x)$ must pass through this cluster center, leveraging the center embedding as the other essential support for determining $\mathcal{P}_i(x)$ besides $A_ix$. Specifically, we employ a learnable $r$-dimensional vector $\boldsymbol{q_i}$ for each expert $i$ as its cluster center, and construct the specific rotation plane $\mathcal{P}_i(x)$ for each embedding $A_ix$ using both $\boldsymbol{q_i}$ and $A_ix$, given by $\mathcal{P}_i(x)=span(\boldsymbol{q_i}, A_ix)$. Considering that $\boldsymbol{q_i}$ and $A_ix$ may not be orthogonal, we treat $\frac{A_ix}{|A_ix|}$ as $\boldsymbol{e_1}$ and the component of $\boldsymbol{q_i}$ orthogonal to $A_ix$ as $\boldsymbol{e_2}$. The whole procedure of constructing the $r$-dimensional rotation matrix $R_i$ can be illustrated by the following equations:

$$ \boldsymbol{e_1}_i = \frac{A_ix}{|A_ix|};$$
$$ \boldsymbol{e_2}_i = \frac{{\boldsymbol{e_2}_i}^*}{|{\boldsymbol{e_2}_i}^*|} \text{, where } {\boldsymbol{e_2}_i}^* = \boldsymbol{q_i}-(\boldsymbol{q_i}\cdot\boldsymbol{e_1}_i)\boldsymbol{e_1}_i;$$

\begin{equation}
\begin{aligned}
R_i= \begin{pmatrix} \boldsymbol{e_1}_i & \boldsymbol{e_2}_i \end{pmatrix} \cdot \begin{pmatrix}
 \cos{\theta_i} & -\sin{\theta_i} \\
 \sin{\theta_i} & \cos{\theta_i}
\end{pmatrix} \cdot \begin{pmatrix}
 {\boldsymbol{e_1}_i}^T \\
 {\boldsymbol{e_2}_i}^T
\end{pmatrix}, \notag
\end{aligned}
\end{equation}

where $g=G(x)$ and $\theta=\Theta(x)$. Here $W_g$, $W_\theta$ and $\boldsymbol{q}$ are learnable parameters, totally $2dn+rn$. Basically, these equations convert a $2$-dimensional rotation matrix into its representation within the $r$-dimensional space via two planar basis vectors, and subsequently apply it to the $r$-dimensional inputs.


\section{Experiments}
\label{exp_big_section}

We conduct a series of comparative experiments to evaluate the effectiveness of RotMoLE across various complex scenarios, including multilingual and multi-task learning. To ensure fine-tuning complexity, we adopt a joint learning manner for all experiments, which means we jointly train on the mixture of all languages or tasks, named as Multilingual or Multi-task Joint Learning (MJL). We compare our RotMoLE with the conventional MoE-LoRA baseline which implements only a scaling gate $g=G(x)$, as well as some recent MoE-LoRA variants with specific structural designs. 

Considering that RotMoLE involves more trainable parameters than the baseline, such as the rotation gate $\theta=\Theta(x)$ or the center vectors $\boldsymbol{q}$, we also perform a comparative experiment for ablation study between our RotMoLE and a size-equivalent MoE-LoRA with an MLP gate, to further illustrate the pure effectiveness of our structural design; Another ablation study we conduct is to verify RotMoLE under different model sizes. The two ablation studies are presented in Appendix \ref{ablation}.

\subsection{Overall Experimental Setup}
\label{overall_exp_setup}

Our proposed method is specifically designed for two circumstances: (1) LoRA expert ranks equal to 2, where we implement only a rotation gate; and (2) LoRA expert ranks larger than 2, where we implement a rotation gate along with $n$ $r$-dimensional trainable vectors. Accordingly, our experiments cover both settings. Besides, we also conduct experiments under both dense MoE and sparse MoE configurations, demonstrating whether our proposed RotMoLE is influenced by the top-$k$ sparse activation. Here we denote the total number of LoRA experts, the number of top experts selected during each forwarding, and the per-expert rank as $n$, $k$, $r$ respectively. To simulate a resource-constrained scenario where experts are shared among multiple tasks, we explicitly constrain that $\frac{n}{k} < N_{task}$, where $N_{task}$ denotes the number of tasks or languages we jointly train. We claim that our proposed RotMoLE provides extra diversity to those task-shared experts by further distinguishing different assigned tasks through rotation operations, and therefore can enhance the overall performance of MJL. For more details of experimental setup, please refer to Appendix \ref{apdx_exp_dtl}.

\subsection{Multi-task Joint Learning}
\subsubsection{Mixture of QA Tasks}
\label{qa_section}

We mix three Question Answering tasks from various fields to constitute a mixture of Multiple QA tasks for our experiments, including CommonsenseQA~\citep{talmor2019commonsenseqa} (CSQA), OpenBookQA~\citep{mihaylov2018can} (OBQA), and SocialIQA~\citep{sap2019socialiqa} (SIQA). Please refer to Appendix \ref{qa_intro} for their detailed introduction. To cover the two circumstances that LoRA ranks both equal to and larger than $2$, we respectively set rank $r$ to $2$ and $3$ in our experiments. We also implement both dense MoE and sparse MoE settings, which consist of $2$ experts with top-$2$ experts activated, and $4$ experts with top-$2$ experts activated, respectively. We train the mixture on Llama-3.2-3B for around 4,000 steps, and on Qwen-2.5-3B for around 5,500 steps, ensuring they are fully updated. The batch size is 135, meaning we train 45 samples for each QA task per step.

We present our overall performances in Table \ref{multiqa_table}, along with detailed per-task performances and the relative standard deviations (Std./Avg.) across tasks. The evaluation metric for QA is accuracy. It is observed that: (1) Our RotMoLE exhibits consistent superiority over the MoE-LoRA baseline (MoE-LoRA) across almost all configurations. Specifically, out of 8 experimental settings, RotMoLE achieves both higher average and per-task accuracy in 7 cases, with only one exception in configuration $\{r=3, n=4, k=2\}$ under Llama-3.2-3B for OpenBookQA task; (2) Quantitatively, RotMoLE can roughly enhance the jointly learning performance on mixed QA tasks by about 1\% $\sim$ 10\%, averagely 5\%; (3) The task-balancing performance of RotMoLE is overall equivalent or slightly better than that of the baseline, according to their relative standard deviation values across tasks; and (4) In most cases except $r=2$ under Qwen-2.5-3B, the performance improvement of RotMoLE over the baseline is more pronounced when the number of experts is very limited, such as $n=2$. This phenomenon indicates the effectiveness of our rotation gate mechanism in boosting per-expert diversity of MoE, since a limited number of experts is more likely to lead to task-shared experts which benefit more from our diversity-boosting design, as we may distinguish tasks by rotation within the same expert. Appendix \ref{convergence_qa} also presents the convergence performances of this experiment. 

\begin{table}[ht]
\centering
\caption{Performance of Jointly Fine-Tuning on Mixed QA Tasks}
\resizebox{\linewidth}{!}{
\begin{tabular}{ccclrrrrrr}
\toprule
                                            &                                          &                                     &                                   &                                          &                                       &                                       &                                    &                                            &                                       \\
\multirow{-2}{*}{\textbf{Base Model}}       & \multirow{-2}{*}{\textbf{$r$}}           & \multirow{-2}{*}{\textbf{$n$, $k$}} & \multirow{-2}{*}{\textbf{Method}} & \multirow{-2}{*}{\textbf{CSQA}}          & \multirow{-2}{*}{\textbf{OBQA}}       & \multirow{-2}{*}{\textbf{SIQA}}       & \multirow{-2}{*}{\textbf{Avg.}}    & \multirow{-2}{*}{\textbf{Boost $\uparrow$}}  & \multirow{-2}{*}{\textbf{Std./Avg.}}   \\ \midrule
                                            &                                          &                                     & MoE-LoRA                          & { 34.73}             & { 42.60}          & { 42.48}          & { 39.93}          & &   0.09  \\
                                            &                                          & \multirow{-2}{*}{2, 2}              & RotMoLE                    & { \textbf{40.21}}    & { \textbf{43.80}} & { \textbf{43.86}} & { \textbf{42.62}} & \multirow{-2}{*}{6.74\%} &   \textbf{0.04}  \\ \cmidrule{3-10} 
                                            &                                          &                                     & MoE-LoRA                          & { 36.94}             & { 40.80}          & { 41.15}          & { 39.63}          & &   0.05  \\
                                            & \multirow{-4}{*}{2}                      & \multirow{-2}{*}{4, 2}              & RotMoLE                    & { \textbf{38.90}}    & { \textbf{41.80}} & { \textbf{43.91}} & { \textbf{41.54}} & \multirow{-2}{*}{4.82\%} &   0.05  \\ \cmidrule{2-10} 
                                            &                                          &                                     & MoE-LoRA                          & { 38.57}             & { 37.00}          & { 40.43}          & { 38.67}          & &   0.04  \\
                                            &                                          & \multirow{-2}{*}{2, 2}              & RotMoLE                    & { \textbf{43.57}}    & { \textbf{41.80}} & { \textbf{42.27}} & { \textbf{42.55}} & \multirow{-2}{*}{10.03\%} &   \textbf{0.02}  \\ \cmidrule{3-10} 
                                            &                                          &                                     & MoE-LoRA                          & { 35.95}             & { \textbf{41.20}} & { 36.34}          & { 37.83}          & &   0.06  \\
\multirow{-8}{*}{Llama-3.2-3B}              & \multirow{-4}{*}{3}                      & \multirow{-2}{*}{4, 2}              & RotMoLE                    & { \textbf{37.67}}    & { 38.20}          & { \textbf{42.27}} & { \textbf{39.38}} & \multirow{-2}{*}{4.10\%} &   \textbf{0.05}  \\ \midrule
                                            &                                          &                                     & MoE-LoRA                          & { 61.18}             & { 56.40}          & { 56.09}          & { 57.89}          & &   \textbf{0.04}  \\
                                            &                                          & \multirow{-2}{*}{2, 2}              & RotMoLE                    & { \textbf{64.13}}    & { \textbf{56.80}} & { \textbf{60.75}} & { \textbf{60.56}} & \multirow{-2}{*}{4.61\%} &   0.05  \\ \cmidrule{3-10} 
                                            &                                          &                                     & MoE-LoRA                          & { 60.20}             & { 52.60}          & { 55.48}          & { 56.09}          & &   0.06  \\
                                            & \multirow{-4}{*}{2}                      & \multirow{-2}{*}{4, 2}              & RotMoLE                    & { \textbf{64.05}}    & { \textbf{60.40}} & { \textbf{60.85}} & { \textbf{61.77}} & \multirow{-2}{*}{10.13\%} &   \textbf{0.03}  \\ \cmidrule{2-10} 
                                            &                                          &                                     & MoE-LoRA                          & { 59.62}             & { 56.20}          & { 57.22}          & { 57.68}          & &   \textbf{0.02}  \\
                                            &                                          & \multirow{-2}{*}{2, 2}              & RotMoLE                    & { \textbf{62.24}}    & { \textbf{57.60}} & { \textbf{58.14}} & { \textbf{59.33}} & \multirow{-2}{*}{2.86\%} &   0.03  \\ \cmidrule{3-10} 
                                            &                                          &                                     & MoE-LoRA                          & { 63.06}             & { 57.00}          & { 56.65}          & { 58.91}          & &   0.05  \\
\multirow{-8}{*}{Qwen-2.5-3B}               & \multirow{-4}{*}{3}                      & \multirow{-2}{*}{4, 2}              & RotMoLE                    & { \textbf{63.88}}    & { \textbf{57.60}} & { \textbf{57.63}} & { \textbf{59.70}} & \multirow{-2}{*}{1.34\%} &   0.05  \\ \bottomrule
\end{tabular}
}
\label{multiqa_table}
\end{table}

\subsubsection{Mixture of GLUE Tasks}
\label{glue_section}

We also mix general NLU tasks from the GLUE benchmark~\citep{wang2018glue} for MJL evaluation. To simplify our experiments, we select only three out of the nine tasks in GLUE: SST-2, MRPC, and QNLI. Details of those tasks are described in Appendix \ref{glue_intro}. Here, we directly set up a sparse MoE configuration and a LoRA rank larger than 2, as they are more general-purpose and are more commonly used. Specifically, we set the rank $r$ of each LoRA expert to 4, and implement two sparse
\begin{wraptable}[11]{r}{0.6\textwidth}
\centering
\caption{Performance of Jointly Fine-Tuning on Mixed GLUE Tasks (Llama-3.2-3B, $r=4$).}
\resizebox{\linewidth}{!}{
\begin{tabular}{clrrrr}
\toprule
                                    &                                   &                                       &                                       &                                       &                                       \\
\multirow{-2}{*}{\textbf{$n$, $k$}} & \multirow{-2}{*}{\textbf{Method}} & \multirow{-2}{*}{\textbf{SST-2}}      & \multirow{-2}{*}{\textbf{MRPC}}       & \multirow{-2}{*}{\textbf{QNLI}}       & \multirow{-2}{*}{\textbf{Avg.}}    \\ \midrule
                                    & MoE-LoRA                          & { 89.33}          & { \textbf{71.77}} & { 66.52}          & { 75.87}          \\
\multirow{-2}{*}{3, 2}              & RotMoLE                    & { \textbf{90.37}} & { 71.07}          & { \textbf{66.67}} & { \textbf{76.03}} \\ \midrule
                                    & MoE-LoRA                          & 88.76                                 & 69.19                                 & 61.63                                 & 73.19                                 \\
\multirow{-2}{*}{5, 3}              & RotMoLE                    & { \textbf{89.22}} & { \textbf{69.21}} & { \textbf{61.83}} & { \textbf{73.42}} \\ \bottomrule
\end{tabular}
}
\label{glue_table}
\end{wraptable}
configurations, $\{n=3, k=2\}$ and $\{n=5, k=3\}$. Both the configurations still follow $\frac{n}{k} < N_{task}$ to model a resource-constrained scenario. We train the mixture of the three tasks on Llama-3.2-3B for around 5,000 steps. The batch size is 30, namely 10 per task. Table \ref{glue_table} illustrates the overall and per-task results. 
Our proposed RotMoLE still exhibits a visible outperformance over MoE-LoRA baseline. 

\subsection{Multilingual Joint Learning}

\subsubsection{Multilingual Title Generation}
\label{mtg_section}

\begin{figure}[h]
\centering
\includegraphics[width=0.8\linewidth]{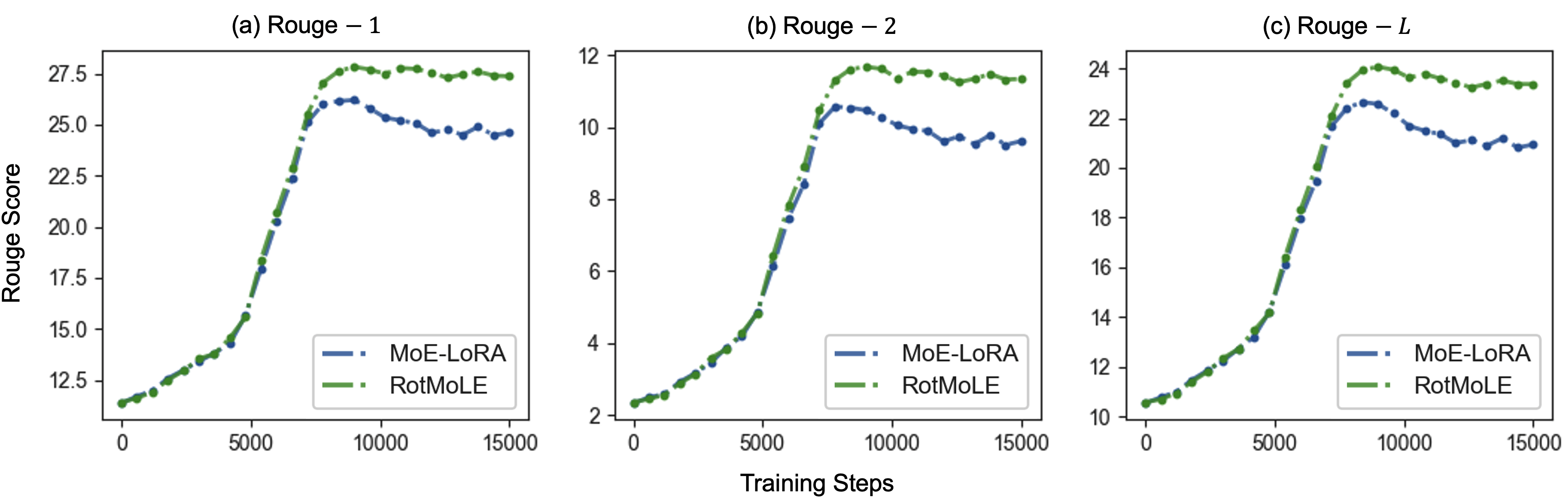}
\caption{Convergence of Jointly Fine-Tuning on Mixed Multilingual Title Generation(Llama-3.2-3B)} 
\label{figure3}
\end{figure}

For multilingual joint learning, we adopt the MTG (Multilingual Text Generation) benchmark proposed by \citet{chen2022mtg}, and take its multilingual title generation sub-task as one of our evaluations. Five languages -- English, Chinese, Spanish, French, and German -- are involved in this task (see Appendix \ref{mtg_intro} for details). We test three MoE configurations: $\{r=3, n=2, k=2\}$, $\{r=2, n=4, k=2\}$, and $\{r=4, n=5, k=3\}$. We train the language mixture on Llama-3.2-3B for 16,000 steps using a batch size of 50 (namely 10 per language). Results are presented in Table \ref{mtg_table}. Here we measure the average performances by Rouge-1, Rouge-2, and Rouge-L scores~\citep{lin2004rouge}.  
To reveal per-language efficacy, we also exhibit Rouge-L scores for each language. In general, RotMoLE still achieves better performances than the MoE-LoRA baseline in terms of both average and per-language Rouge metrics in most cases.  
We also compare their convergence performances under one of the MoE configurations: $\{r=4, n=5, k=3\}$ in Figure \ref{figure3}. It shows that both methods initially update equivalently during the first 8,000 steps, but the MoE-LoRA baseline exhibits an earlier convergence, and overfits much faster than our proposed RotMoLE. This further indicates that our rotation design may introduce some extra robustness to model against routers overfitting.

\begin{table}[h]
\centering
\caption{Performance of Jointly Fine-Tuning on Mixed Multilingual Title Generation (Llama-3.2-3B).} 
\resizebox{\linewidth}{!}{
\begin{tabular}{clrrrrrrrr}
\toprule
                                &                                   & \textbf{EN}                           & \textbf{ZH}                           & \textbf{ES}                           & \textbf{FR}                           & \textbf{DE}                           & \multicolumn{3}{c}{\textbf{Avg.}}                                                                                  \\ \cmidrule{3-10} 
\multirow{-2}{*}{$r$, $n$, $k$} & \multirow{-2}{*}{\textbf{Method}} & \multicolumn{5}{c}{\textbf{Rouge-L}}                                                                                                                                                                  & \textbf{Rouge-1}                      & \textbf{Rouge-2}                      & \textbf{Rouge-L}                      \\ \midrule
                                & MoE-LoRA                          & { 32.33}          & { 23.87}          & { 27.80}          & { 25.28}          & { 22.79}          & { 29.63}          & { 11.76}          & { 26.41}          \\
\multirow{-2}{*}{3, 2, 2}       & RotMoLE                    & { \textbf{32.92}} & { \textbf{24.35}} & { \textbf{28.02}} & { \textbf{25.37}} & { \textbf{23.17}} & { \textbf{30.21}} & { \textbf{12.50}} & { \textbf{26.76}} \\ \midrule
                                & MoE-LoRA                          & { 30.14}          & { 26.94}          & { 28.04}          & { \textbf{27.00}}          & { 23.11}          & { 30.26}          & { 11.67}          & { 27.04}          \\
\multirow{-2}{*}{2, 4, 2}       & RotMoLE                    & { \textbf{32.79}} & { \textbf{26.96}} & { \textbf{28.72}} & { 26.85} & { \textbf{23.39}} & { \textbf{30.68}} & { \textbf{12.48}} & { \textbf{27.74}} \\ \midrule
                                & MoE-LoRA                          & { 23.40}          & { 24.27}          & { 24.76}          & { 22.73}          & { 18.07}          & { 26.14}          & { 10.54}          & { 22.64}          \\
\multirow{-2}{*}{4, 5, 3}       & RotMoLE                    & { \textbf{25.28}} & { \textbf{25.48}} & { \textbf{25.75}} & { \textbf{24.90}} & { \textbf{18.89}} & { \textbf{27.82}} & { \textbf{11.67}} & { \textbf{24.06}} \\ \bottomrule
\end{tabular}
}
\label{mtg_table}
\end{table}

\subsubsection{Multilingual NLI}
\label{xnli_section}

Here we adopt XNLI~\citep{conneau2018xnli} for our multilingual natural language inference experiments. Appendix \ref{xnli_intro} describes the details of XNLI corpus. Notably, the original purpose of XNLI is to evaluate the cross-lingual transfer ability of multilingual models (i.e., whether models purely trained on English NLI data can handle NLI in other languages). However, in this paper we treat XNLI as a multilingual joint learning dataset to evaluate the performances of models jointly trained on the mixture of samples in five languages. 
Here we directly set up a sparse MoE and a LoRA expert rank larger than 2, as they are more common scenarios, that is $\{r=4, n=5, k=3\}$. We jointly train the mixture on Llama-3.2-3B for around 4,500 steps until they are fully updated, using a batch size of 50 with 10 per language. The overall and per-language accuracies are illustrated in Table \ref{xnli_table}. Besides the overall performance, our proposed RotMoLE also outperforms the MoE-LoRA baseline in four out of the five languages. This also indicates that RotMoLE learns more equivalent capacities across different languages during the joint training process.   

\begin{table}[h]
\centering
\caption{Performance of Jointly Fine-Tuning on Mixed Multilingual NLI (Llama-3.2-3B).} 
\resizebox{0.75\linewidth}{!}{
\begin{tabular}{clrrrrrr}
\toprule
\textbf{$r$, $n$, $k$} & \textbf{Method} & \textbf{EN}     & \textbf{ZH}       & \textbf{ES}       & \textbf{FR}       & \textbf{DE}      & \textbf{Avg.}     \\ \midrule
 & MoE-LoRA        & { \textbf{52.25}} & { 46.30}          & { 48.85}          & { 48.90}          & { 47.45}          & { 48.75}          \\
\multirow{-2}{*}{4, 5, 3} & RotMoLE  & { 51.96}          & { \textbf{47.94}} & { \textbf{50.38}} & { \textbf{49.14}} & { \textbf{49.12}} & { \textbf{49.71}} \\ \bottomrule
\end{tabular}
}
\label{xnli_table}
\end{table}

\subsection{Comparison with Other Baselines}
\label{baseline_section}

Although there are many variants proposed for enhancing MoE-LoRA, most of them focus on the loss design during back-propagation, such as the load-balancing loss or the contrastive loss. Considering that our proposed RotMoLE is a gating-based forward method, here we compare our RotMoLE with three variants of MoE-LoRA that specifically focus on improving forward routing effectiveness, including MoCLE~\citep{gou2023mixture}, MoLA~\citep{mola}, and AdaMoLE~\citep{liu2024adamole}. MoCLE first clusters the input embeddings into several clusters, and then conducts a top-$1$ cluster-level expert routing along with a permanently-activated universal expert; MoLA assigns more experts to higher layers given a fixed total number of experts; AdaMoLE replaces the top-$k$ activation by a top-$p$ activation which selects experts whose gate values are larger than a threshold $p$, and dynamically adjusts $p$ by a network. 

We conduct our comparisons on the mixture of QA tasks in Section \ref{qa_section}, and apply Qwen-2.5-3B as our foundation model. Since MoCLE implements a top-$1$ activation and an extra universal expert, it is distinctive only under the condition of sparse MoE with 2 experts activated. As a result, we set the number of experts $n$ to 4, and the number of activated experts $k$ to 2. Additionally, we naturally set the number of clusters in MoCLE to the number of distinct tasks jointly trained during our experiments, which is 3; For MoLA, we set the expert allocation strategy among the first nine layers of
\begin{wraptable}[11]{r}{0.6\textwidth}
\centering
\caption{Performance of Jointly Fine-Tuning on Mixed QA Tasks by Various MoE-LoRA Methods (Qwen-2.5-3B, $r=6,n=4,k=2$).} 
\begin{tabular}{lrrrr}
\toprule
\textbf{Method} & \multicolumn{1}{r}{\textbf{CSQA}}                             & \multicolumn{1}{r}{\textbf{OBQA}}                             & \multicolumn{1}{r}{\textbf{SIQA}}                             & \textbf{Avg.}                                                 \\ \midrule
MoCLE           & { 36.28}          & { 29.60}          & { 37.82}          & { 34.57}          \\
MoLA            & 39.31                                 & 30.00                                 & 37.67                                 & 35.66                                 \\
AdaMoLE         & \textbf{49.39}                                 & 37.00                                 & 37.87                                 & 41.42                                 \\
RotMoLE (Ours)  & { 43.00} & { \textbf{46.60}} & { \textbf{47.85}} & { \textbf{45.82}} \\ \bottomrule
\end{tabular}
\label{baseline_compare}
\end{wraptable}
Qwen-2.5-3B to $[2,2,2,4,4,4,6,6,$ $6]$, which is equivalent to the configuration of 4 experts per layer in terms of total expert count. Finally, we set the per-expert rank $r$ to 6 to directly test our method under the circumstance of large ranks. Table \ref{baseline_compare} shows the results.  
An outperformance of our RotMoLE over the other three MoE-LoRA variants is observed, in terms of both overall and most of per-task accuracies.  

\subsection{Rotation Distribution Analysis}
\label{ana_section}

\begin{figure}[h]
\centering
\includegraphics[width=\linewidth]{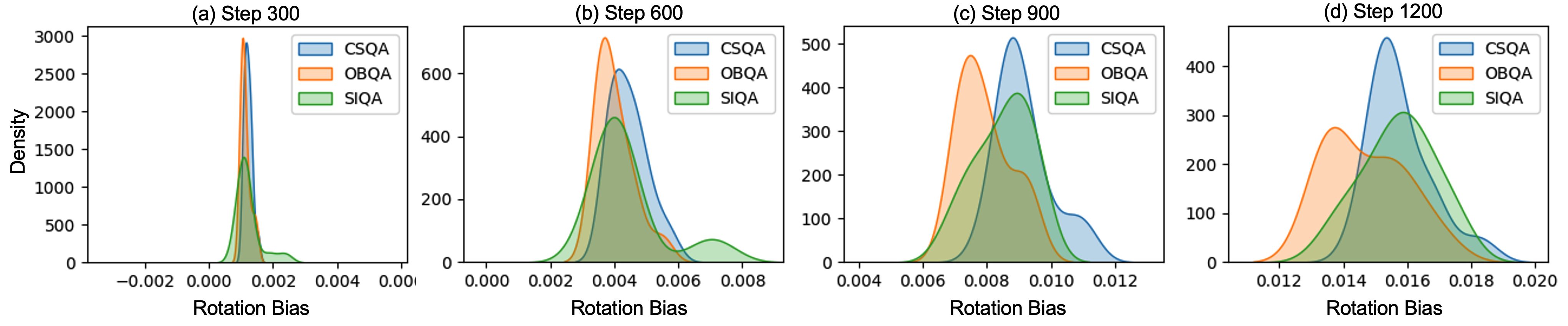}
\caption{Rotation bias distribution of the first expert within the $v\_proj$ module in the 8-th layer. }
\label{figure_distribution}
\end{figure}

We also conduct internal probing experiments for our proposed RotMoLE, to analyze the distribution of rotational biases (angles) allocated to different sub-tasks. Specifically, during our validating experiments on the mixture of QA tasks under Qwen-2.5-3B, we store the per-task rotation bias value $\theta$s for the first expert within the $v\_proj$ module in the 8-th layer (the last layer with trainable MoE-LoRA adapters). Then we compute the sequence-level rotation bias by averaging the rotation bias values for each token within every validating sequence. Finally, we plot the distribution of sequence-level rotation biases for each sub-task. Figure \ref{figure_distribution} exhibits this distribution after 300, 600, 900, and 1,200 training steps, respectively. Here the configuration is $\{r=2, n=4, k=2\}$. It is observed that: (1) During training process, rotation biases become larger and non-zero, and their distributions become more flat and dispersed, indicating our rotations are playing roles in modeling sample diversity, aligning with our illustrations in Figure \ref{figure1};  
and (2) The deviation between three sub-task distributions becomes more obvious, indicating our rotations are also playing roles in sub-task specialization and modeling the three QA sub-tasks in different ways.

\section{Conclusion}

We propose the Rotatable Mixture of Low-rank Experts (i.e., RotMoLE), which is a novel MoE-LoRA framework that enhances the representational capacity of conventional MoE-LoRA in diverse and complicated scenarios through an additional rotational gating mechanism. Unlike conventional MoE approaches that rely solely on scalar scaling operations for expert aggregation, RotMoLE introduces a rotation gate that performs 2D rotations on $r$-dimensional low-rank space of experts, thereby significantly expanding the solution space of weighted expert combinations. 
Our extensive experiments across multi-task and multilingual joint learning demonstrate the effectiveness and generalization ability of RotMoLE. Appendix \ref{limitations} discusses the limitations.


\bibliographystyle{plainnat}
\bibliography{reference}

@article{jacobs1991adaptive,
  title={{Adaptive Mixtures of Local Experts}},
  author={Jacobs, Robert A and Jordan, Michael I and Nowlan, Steven J and Hinton, Geoffrey E},
  journal={Neural Computation},
  volume={3},
  number={1},
  pages={79--87},
  year={1991},
  publisher={MIT Press}
}

@article{cai2025survey,
  title={{A Survey on Mixture of Experts in Large Language Models}},
  author={Cai, Weilin and Jiang, Juyong and Wang, Fan and Tang, Jing and Kim, Sunghun and Huang, Jiayi},
  journal={IEEE Transactions on Knowledge and Data Engineering},
  year={2025},
  publisher={IEEE}
}

@article{shazeer2017outrageously,
  title={{Outrageously Large Neural Networks: The Sparsely-Gated Mixture-of-Experts Layer}},
  author={Shazeer, Noam and Mirhoseini, Azalia and Maziarz, Krzysztof and Davis, Andy and Le, Quoc and Hinton, Geoffrey and Dean, Jeff},
  journal={International Conference on Learning Representations},
  year={2017}
}

@article{ren2023pangu,
  title={{PanGu-$\{$$\backslash$Sigma$\}$: Towards Trillion Parameter Language Model with Sparse Heterogeneous Computing}},
  author={Ren, Xiaozhe and Zhou, Pingyi and Meng, Xinfan and Huang, Xinjing and Wang, Yadao and Wang, Weichao and others},
  journal={arXiv e-prints},
  pages={arXiv--2303},
  year={2023}
}

@article{zhang2025mixture,
  title={{Mixture of Experts in Large Language Models}},
  author={Zhang, Danyang and Song, Junhao and Bi, Ziqian and Yuan, Yingfang and Wang, Tianyang and Yeong, Joe and Hao, Junfeng},
  journal={arXiv e-prints},
  pages={arXiv--2507},
  year={2025}
}

@article{nie2021dense,
  title={{EvoMoE: An Evolutional Mixture-of-Experts Training Framework via Dense-To-Sparse Gate}},
  author={Nie, Xiaonan and Miao, Xupeng and Cao, Shijie and Ma, Lingxiao and Liu, Qibin and Xue, Jilong and Miao, Youshan and Liu, Yi and Yang, Zhi and Cui, Bin},
  journal={arXiv e-prints},
  pages={arXiv--2112},
  year={2021}
}

@article{zhang2025parameter,
  title={{Parameter-Efficient Fine-Tuning for Foundation Models}},
  author={Zhang, Dan and Feng, Tao and Xue, Lilong and Wang, Yuandong and Dong, Yuxiao and Tang, Jie},
  journal={arXiv e-prints},
  pages={arXiv--2501},
  year={2025}
}

@article{hu2022lora,
  title={{LoRA: Low-Rank Adaptation of Large Language Models}},
  author={Hu, Edward J and Shen, Yelong and Wallis, Phillip and Allen-Zhu, Zeyuan and Li, Yuanzhi and Wang, Shean and others},
  journal={International Conference on Learning Representations},
  volume={1},
  number={2},
  pages={3},
  year={2022}
}

@article{dou2024loramoe,
  title={{LoRAMoE: Alleviating World Knowledge Forgetting in Large Language Models via MoE-Style Plugin}},
  author={Dou, Shihan and Zhou, Enyu and Liu, Yan and Gao, Songyang and Shen, Wei and Xiong, Limao and others},
  journal={Proceedings of the 62nd Annual Meeting of the Association for Computational Linguistics (Volume 1: Long Papers)},
  pages={1932--1945},
  year={2024}
}

@article{li2024mixlora,
  title={{MixLoRA: Enhancing Large Language Models Fine-Tuning with LoRA-based Mixture of Experts}},
  author={Li, Dengchun and Ma, Yingzi and Wang, Naizheng and Ye, Zhengmao and Cheng, Zhiyuan and Tang, Yinghao and others},
  journal={arXiv e-prints},
  pages={arXiv--2404},
  year={2024}
}

@article{liu2023moelora,
  title={{When MoE Meets LLMs: Parameter Efficient Fine-Tuning for Multi-Task Medical Applications}},
  author={Liu, Qidong and Wu, Xian and Zhao, Xiangyu and Zhu, Yuanshao and Xu, Derong and Tian, Feng and Zheng, Yefeng},
  journal={Proceedings of the 47th International ACM SIGIR Conference on Research and Development in Information Retrieval},
  pages={1104--1114},
  year={2024}
}

@article{gou2023mixture,
  title={{Mixture of Cluster-Conditional LoRA Experts for Vision-Language Instruction Tuning}},
  author={Gou, Yunhao and Liu, Zhili and Chen, Kai and Hong, Lanqing and Xu, Hang and Li, Aoxue and Yeung, Dit-Yan and Kwok, James T and Zhang, Yu},
  journal={arXiv e-prints},
  pages={arXiv--2312},
  year={2023}
}

@article{gedynamic,
  title={{Dynamic Mixture of Curriculum LoRA Experts for Continual Multimodal Instruction Tuning}},
  author={Ge, Chendi and Wang, Xin and Zhang, Zeyang and Chen, Hong and Fan, Jiapei and Huang, Longtao and Xue, Hui and Zhu, Wenwu},
  journal={Forty-second International Conference on Machine Learning},
  year={2025}
}

@article{zhao2023survey,
  title={{A Survey of Large Language Models}},
  author={Zhao, Wayne Xin and Zhou, Kun and Li, Junyi and Tang, Tianyi and Wang, Xiaolei and Hou, Yupeng and Min, Yingqian and others},
  journal={arXiv e-prints},
  pages={arXiv--2303},
  year={2023}
}

@article{collobert2001parallel,
  title={{A Parallel Mixture of SVMs for Very Large Scale Problems}},
  author={Collobert, Ronan and Bengio, Samy and Bengio, Yoshua},
  journal={Advances in Neural Information Processing Systems},
  volume={14},
  year={2001}
}

@article{eigen2013learning,
  title={{Learning Factored Representations in a Deep Mixture of Experts}},
  author={Eigen, David and Ranzato, Marc'Aurelio and Sutskever, Ilya},
  journal={arXiv e-prints},
  pages={arXiv--1312},
  year={2013}
}

@article{xu1994alternative,
  title={{An Alternative Model for Mixtures of Experts}},
  author={Xu, Lei and Jordan, Michael and Hinton, Geoffrey E},
  journal={Advances in Neural Information Processing Systems},
  volume={7},
  year={1994}
}

@article{yuksel2012twenty,
  title={{Twenty Years of Mixture of Experts}},
  author={Yuksel, Seniha Esen and Wilson, Joseph N and Gader, Paul D},
  journal={IEEE Transactions on Neural Networks and Learning Systems},
  volume={23},
  number={8},
  pages={1177--1193},
  year={2012},
  publisher={IEEE}
}

@article{lepikhin2020gshard,
  title={{GShard: Scaling Giant Models with Conditional Computation and Automatic Sharding}},
  author={Lepikhin, Dmitry and Lee, HyoukJoong and Xu, Yuanzhong and Chen, Dehao and Firat, Orhan and Huang, Yanping and Krikun, Maxim and Shazeer, Noam and Chen, Zhifeng},
  journal={International Conference on Learning Representations},
  year={2021}
}

@article{touvron2023llama,
  title={{LLaMA: Open and Efficient Foundation Language Models}},
  author={Touvron, Hugo and Lavril, Thibaut and Izacard, Gautier and Martinet, Xavier and Lachaux, Marie-Anne and others},
  journal={arXiv e-prints},
  pages={arXiv--2302},
  year={2023}
}

@article{liu2024deepseek,
  title={{DeepSeek-V3 Technical Report}},
  author={Liu, Aixin and Feng, Bei and Xue, Bing and Wang, Bingxuan and Wu, Bochao and Lu, Chengda and Zhao, Chenggang and others},
  journal={arXiv e-prints},
  pages={arXiv--2412},
  year={2024}
}

@article{jiang2024mixtral,
  title={{Mixtral of Experts}},
  author={Jiang, Albert Q and Sablayrolles, Alexandre and Roux, Antoine and Mensch, Arthur and Savary, Blanche and others},
  journal={arXiv e-prints},
  pages={arXiv--2401},
  year={2024}
}

@article{liu2022few,
  title={{Few-Shot Parameter-Efficient Fine-Tuning is Better and Cheaper Than In-Context Learning}},
  author={Liu, Haokun and Tam, Derek and Muqeeth, Mohammed and Mohta, Jay and Huang, Tenghao and Bansal, Mohit and Raffel, Colin A},
  journal={Advances in Neural Information Processing Systems},
  volume={35},
  pages={1950--1965},
  year={2022}
}

@article{lester2021power,
  title={{The Power of Scale For Parameter-Efficient Prompt Tuning}},
  author={Lester, Brian and Al-Rfou, Rami and Constant, Noah},
  journal={Proceedings of the 2021 Conference on Empirical Methods in Natural Language Processing},
  pages={3045--3059},
  year={2021}
}

@article{li2021prefix,
  title={{Prefix-Tuning: Optimizing Continuous Prompts for Generation}},
  author={Li, Xiang Lisa and Liang, Percy},
  journal={Proceedings of the 59th Annual Meeting of the Association for Computational Linguistics and the 11th International Joint Conference on Natural Language Processing (Volume 1: Long Papers)},
  pages={4582--4597},
  year={2021}
}

@article{liu2022p,
  title={{P-Tuning: Prompt Tuning Can Be Comparable to Fine-Tuning Across Scales and Tasks}},
  author={Liu, Xiao and Ji, Kaixuan and Fu, Yicheng and Tam, Weng and Du, Zhengxiao and Yang, Zhilin and Tang, Jie},
  journal={Proceedings of the 60th Annual Meeting of the Association for Computational Linguistics (Volume 2: Short Papers)},
  pages={61--68},
  year={2022}
}

@article{han2024parameter,
  title={{Parameter-Efficient Fine-Tuning for Large Models: A Comprehensive Survey}},
  author={Zeyu Han and Chao Gao and Jinyang Liu and Jeff Zhang and Sai Qian Zhang},
  journal={Transactions on Machine Learning Research},
  issn={2835-8856},
  year={2024}
}

@article{zhang2025optimizing,
  title={{Optimizing Robustness and Accuracy in Mixture of Experts: A Dual-Model Approach}},
  author={Zhang, Xu and Xu, Kaidi and Hu, Ziqing and Wang, Ren},
  journal={International Conference on Machine Learning},
  pages={76436--76450},
  year={2025},
  organization={PMLR}
}

@article{ma2024big,
  title={{BIG-MoE: Bypass Isolated Gating MoE for Generalized Multimodal Face Anti-Spoofing}},
  author={Ma, Yingjie and Yu, Zitong and Lin, Xun and Xie, Weicheng and Shen, Linlin},
  journal={arXiv e-prints},
  pages={arXiv--2412},
  year={2024}
}

@article{eo2025mixture,
  title={{Mixture-of-Clustered-Experts: Advancing Expert Specialization and Generalization in Instruction Tuning}},
  author={Eo, Sugyeong and Lee, Jung Jun and Park, Chanjun and Lim, Heui-Seok},
  journal={Proceedings of the 2025 Conference on Empirical Methods in Natural Language Processing},
  pages={14212--14223},
  year={2025}
}

@article{guo2024dynamic,
  title={{Dynamic Mixture of Experts: An Auto-Tuning Approach for Efficient Transformer Models}},
  author={Guo, Yongxin and Cheng, Zhenglin and Tang, Xiaoying and Tu, Zhaopeng and Lin, Tao},
  journal={The Thirteenth International Conference on Learning Representations},
  year={2025}
}

@article{zhou2022mixture,
  title={{Mixture-of-Experts with Expert Choice Routing}},
  author={Zhou, Yanqi and Lei, Tao and Liu, Hanxiao and Du, Nan and Huang, Yanping and Zhao, Vincent and others},
  journal={Advances in Neural Information Processing Systems},
  volume={35},
  pages={7103--7114},
  year={2022}
}

@article{wu2024yuan,
  title={{Yuan 2.0-M32: Mixture of Experts with Attention Router}},
  author={Wu, Shaohua and Luo, Jiangang and Chen, Xi and Li, Lingjun and Zhao, Xudong and Yu, Tong and Wang, Chao and others},
  journal={arXiv e-prints},
  pages={arXiv--2405},
  year={2024}
}

@article{harvey2025optimizing,
  title={{Optimizing MoE Routers: Design, Implementation, and Evaluation in Transformer Models}},
  author={Harvey, Daniel Fidel and Weale, George and Yilmaz, Berk},
  journal={arXiv preprint arXiv:2506.16419},
  year={2025}
}

@article{luo2024moelora,
  title={{MoELoRA: Contrastive Learning Guided Mixture of Experts on Parameter-Efficient Fine-Tuning for Large Language Models}},
  author={Luo, Tongxu and Lei, Jiahe and Lei, Fangyu and Liu, Weihao and He, Shizhu and Zhao, Jun and Liu, Kang},
  journal={arXiv e-prints},
  pages={arXiv--2402},
  year={2024}
}

@article{touvron2023llama2,
  title={{Llama 2: Open Foundation and Fine-Tuned Chat Models}},
  author={Touvron, Hugo and Martin, Louis and Stone, Kevin and Albert, Peter and Almahairi, Amjad and Babaei, Yasmine and others},
  journal={arXiv e-prints},
  pages={arXiv--2307},
  year={2023}
}

@article{qwen2,
  title={{Qwen2.5 Technical Report}},
  author={Yang, An and Yang, Baosong and Zhang, Beichen and Hui, Binyuan and Zheng, Bo and Yu, Bowen and Li, Chengyuan and others},
  journal={arXiv e-prints},
  pages={arXiv--2412},
  year={2024}
}

@article{glm2024chatglm,
  title={{ChatGLM: A Family of Large Language Models from GLM-130B to GLM-4 All Tools}},
  author={Team, GLM and Zeng, Aohan and Xu, Bin and Wang, Bowen and Zhang, Chenhui and Yin, Da and Zhang, Dan and others},
  journal={arXiv e-prints},
  pages={arXiv--2406},
  year={2024}
}

@article{conneau2018xnli,
  title={{XNLI: Evaluating Cross-Lingual Sentence Representations}},
  author={Conneau, Alexis and Rinott, Ruty and Lample, Guillaume and Williams, Adina and Bowman, Samuel and Schwenk, Holger and Stoyanov, Veselin},
  journal={Proceedings of the 2018 Conference on Empirical Methods in Natural Language Processing},
  pages={2475--2485},
  year={2018}
}

@article{talmor2019commonsenseqa,
  title={{CommonsenseQA: A Question Answering Challenge Targeting Commonsense Knowledge}},
  author={Talmor, Alon and Herzig, Jonathan and Lourie, Nicholas and Berant, Jonathan},
  journal={Proceedings of the 2019 Conference of the North American Chapter of the Association for Computational Linguistics: Human Language Technologies, Volume 1 (Long and Short Papers)},
  pages={4149--4158},
  year={2019}
}

@article{mihaylov2018can,
  title={{Can a Suit of Armor Conduct Electricity? A New Dataset for Open Book Question Answering}},
  author={Mihaylov, Todor and Clark, Peter and Khot, Tushar and Sabharwal, Ashish},
  journal={Proceedings of the 2018 Conference on Empirical Methods in Natural Language Processing},
  pages={2381--2391},
  year={2018}
}

@article{sap2019socialiqa,
  title={{Social IQA: Commonsense Reasoning About Social Interactions}},
  author={Sap, Maarten and Rashkin, Hannah and Chen, Derek and Le Bras, Ronan and Choi, Yejin},
  journal={Proceedings of the 2019 Conference on Empirical Methods in Natural Language Processing and the 9th International Joint Conference on Natural Language Processing (EMNLP-IJCNLP)},
  pages={4463--4473},
  year={2019}
}

@article{wang2018glue,
  title={{GLUE: A Multi-Task Benchmark and Analysis Platform for Natural Language Understanding}},
  author={Wang, Alex and Singh, Amanpreet and Michael, Julian and Hill, Felix and Levy, Omer and Bowman, Samuel},
  journal={Proceedings of the 2018 EMNLP Workshop BlackboxNLP: Analyzing and Interpreting Neural Networks for NLP},
  pages={353--355},
  year={2018}
}

@article{rajpurkar2016squad,
  title={{SQuAD: 100,000+ Questions for Machine Comprehension of Text}},
  author={Rajpurkar, Pranav and Zhang, Jian and Lopyrev, Konstantin and Liang, Percy},
  journal={Proceedings of the 2016 Conference on Empirical Methods in Natural Language Processing},
  pages={2383--2392},
  year={2016}
}

@article{chen2022mtg,
  title={{MTG: A Benchmark Suite for Multilingual Text Generation}},
  author={Chen, Yiran and Song, Zhenqiao and Wu, Xianze and Wang, Danqing and Xu, Jingjing and Chen, Jiaze and Zhou, Hao and Li, Lei},
  journal={Findings of the Association for Computational Linguistics: NAACL 2022},
  pages={2508--2527},
  year={2022}
}

@article{lin2004rouge,
  title={{Rouge: A Package for Automatic Evaluation of Summaries}},
  author={Lin, Chin-Yew},
  journal={Text Summarization Branches Out},
  pages={74--81},
  year={2004}
}

@article{pissa,
  title={{PiSSA: Principal Singular Values and Singular Vectors Adaptation of Large Language Models}},
  author={Meng, Fanxu and Wang, Zhaohui and Zhang, Muhan},
  journal={Advances in Neural Information Processing Systems},
  volume={37},
  pages={121038--121072},
  year={2024}
}

@article{milora,
  title={{MiLoRA: Harnessing Minor Singular Components for Parameter-Efficient LLM Finetuning}},
  author={Wang, Hanqing and Li, Yixia and Wang, Shuo and Chen, Guanhua and Chen, Yun},
  journal={Proceedings of the 2025 Conference of the Nations of the Americas Chapter of the Association for Computational Linguistics: Human Language Technologies (Volume 1: Long Papers)},
  pages={4823--4836},
  year={2025}
}

@article{lorapro,
  title={{LoRA-Pro: Are Low-Rank Adapters Properly Optimized?}},
  author={Wang, Zhengbo and Liang, Jian and He, Ran and Wang, Zilei and Tan, Tieniu},
  journal={The Thirteenth International Conference on Learning Representations},
  year={2025}
}

@article{lora-ga,
  title={{LoRA-GA: Low-Rank Adaptation with Gradient Approximation}},
  author={Wang, Shaowen and Yu, Linxi and Li, Jian},
  journal={Advances in Neural Information Processing Systems},
  volume={37},
  pages={54905--54931},
  year={2024}
}

@article{lora+,
  title={{LoRA+: Efficient Low Rank Adaptation of Large Models}},
  author={Hayou, Soufiane and Ghosh, Nikhil and Yu, Bin},
  journal={Proceedings of the 41st International Conference on Machine Learning},
  pages={17783--17806},
  year={2024}
}

@article{dora,
  title={{DoRA: Weight-Decomposed Low-Rank Adaptation}},
  author={Liu, Shih-Yang and Wang, Chien-Yi and Yin, Hongxu and Molchanov, Pavlo and Wang, Yu-Chiang Frank and Cheng, Kwang-Ting and Chen, Min-Hung},
  journal={Forty-first International Conference on Machine Learning},
  year={2024}
}

@article{rslora,
  title={{A Rank Stabilization Scaling Factor for Fine-Tuning with LoRA}},
  author={Kalajdzievski, Damjan},
  journal={arXiv e-prints},
  pages={arXiv--2312},
  year={2023}
}

@article{hiddenkey,
  title={{LoRA Meets Dropout Under A Unified Framework}},
  author={Wang, Sheng and Chen, Liheng and Jiang, Jiyue and Xue, Boyang and Kong, Lingpeng and Wu, Chuan},
  journal={Findings of the Association for Computational Linguistics: ACL 2024},
  pages={1995--2008},
  year={2024}
}

@article{loradropout,
  title={{LoRA Dropout as a Sparsity Regularizer for Overfitting Control}},
  author={Lin, Yang and Ma, Xinyu and Chu, Xu and Jin, Yujie and Yang, Zhibang and Wang, Yasha and Mei, Hong},
  journal={arXiv e-prints},
  pages={arXiv--2404},
  year={2024}
}

@article{zhang2024riemannian,
  title={{Riemannian Preconditioned LoRA for Fine-Tuning Foundation Models}},
  author={Zhang, Fangzhao and Pilanci, Mert},
  journal={Proceedings of the 41st International Conference on Machine Learning},
  pages={59641--59669},
  year={2024}
}

@article{adamix,
  title={{AdaMix: Mixture-of-Adaptations for Parameter-Efficient Model Tuning}},
  author={Wang, Yaqing and Agarwal, Sahaj and Mukherjee, Subhabrata and Liu, Xiaodong and Gao, Jing and Hassan, Ahmed and Gao, Jianfeng},
  journal={Proceedings of the 2022 Conference on Empirical Methods in Natural Language Processing},
  pages={5744--5760},
  year={2022}
}

@article{yang2024moral,
  title={{MoRAL: MoE Augmented LoRA for LLMs' Lifelong Learning}},
  author={Yang, Shu and Ali, Muhammad Asif and Wang, Cheng-Long and Hu, Lijie and Wang, Di},
  journal={arXiv e-prints},
  pages={arXiv--2402},
  year={2024}
}

@article{mola,
  title={{Higher Layers Need More LoRA Experts}},
  author={Gao, Chongyang and Chen, Kezhen and Rao, Jinmeng and Sun, Baochen and Liu, Ruibo and Peng, Daiyi and Zhang, Yawen and Guo, Xiaoyuan and Yang, Jie and Subrahmanian, VS},
  journal={arXiv e-prints},
  pages={arXiv--2402},
  year={2024}
}

@article{sun2025stronger,
  title={{A Stronger Mixture of Low-Rank Experts for Fine-Tuning Foundation Models}},
  author={Sun, Mengyang and Wang, Yihao and Feng, Tao and Zhang, Dan and Zhu, Yifan and Tang, Jie},
  journal={International Conference on Machine Learning},
  pages={57712--57727},
  year={2025},
  organization={PMLR}
}

@article{mole,
  title={{Mixture of LoRA Experts}},
  author={Wu, Xun and Huang, Shaohan and Wei, Furu},
  journal={The Twelfth International Conference on Learning Representations},
  year={2024}
}

@article{sira,
  title={{SiRA: Sparse Mixture of Low Rank Adaptation}},
  author={Zhu, Yun and Wichers, Nevan and Lin, Chu-Cheng and Wang, Xinyi and Chen, Tianlong and Shu, Lei and Lu, Han and others},
  journal={arXiv e-prints},
  pages={arXiv--2311},
  year={2023}
}

@article{liu2024adamole,
  title={{AdaMoLE: Fine-Tuning Large Language Models with Adaptive Mixture of Low-Rank Adaptation Experts}},
  author={Liu, Zefang and Luo, Jiahua},
  journal={First Conference on Language Modeling},
  year={2024}
}

@article{shah2026molora,
  title={{MoLoRA: Composable Specialization via Per-Token Adapter Routing}},
  author={Shah, Shrey and Wagle, Justin},
  journal={arXiv preprint arXiv:2603.15965},
  year={2026}
}

@article{mu2026talklora,
  title={{TalkLoRA: Communication-Aware Mixture of Low-Rank Adaptation for Large Language Models}},
  author={Mu, Lin and Wang, Haiyang and Ni, Li and Sang, Lei and Wu, Zhize and Jin, Peiquan and Zhang, Yiwen},
  journal={arXiv preprint arXiv:2604.06291},
  year={2026}
}

@article{cao2026comol,
  title={{CoMoL: Efficient Mixture of LoRA Experts via Dynamic Core Space Merging}},
  author={Cao, Jie and Fan, Zhenxuan and Wang, Zhuonan and Lin, Tianwei and Zhao, Ziyuan and Yan, Rolan and Zhang, Wenqiao and Shao, Feifei and Wang, Hongwei and Xiao, Jun and others},
  journal={arXiv preprint arXiv:2603.00573},
  year={2026}
}

@article{qiu2026remix,
  title={{ReMix: Reinforcement Routing for Mixtures of LoRAs in LLM Finetuning}},
  author={Qiu, Ruizhong and Zeng, Hanqing and Xia, Yinglong and Meng, Yiwen and Chen, Ren and Feng, Jiarui and Fu, Dongqi and Wang, Qifan and Liu, Jiayi and Xiao, Jun and others},
  journal={arXiv preprint arXiv:2603.10160},
  year={2026}
}

@article{shi2026samora,
  title={{SAMoRA: Semantic-Aware Mixture of LoRA Experts for Task-Adaptive Learning}},
  author={Shi, Boyan and Chen, Wei and Zhao, Shuyuan and Shen, Junfeng and Guo, Shengnan and Wang, Shaojiang and Wan, Huaiyu},
  journal={arXiv preprint arXiv:2604.19048},
  year={2026}
}


\appendix

\section{Related Work}
\label{apdx_related_work}

\subsection{Mixture of Experts in LLMs}

The MoE architecture, known as a framework for efficiently increasing model capacity without significant computational overhead, consists of a series of specialized modules called \emph{experts}, alongside a routing module known as \emph{gate}. In the concept of MoE, each expert is dedicated to a specific domain or task, or granular features of the data~\citep{cai2025survey}. This fundamental idea was initially introduced by \citet{jacobs1991adaptive} to alleviate the issue of cross-interference between tasks and enhance training efficiency. Subsequent studies, such as \citet{xu1994alternative}, \citet{collobert2001parallel}, \citet{yuksel2012twenty}, and \citet{eigen2013learning}, have all made impressive contributions to the field of MoE. Depending on how the gating module is implemented, MoE architectures can be basically separated into two types: dense MoE and sparse MoE. 

Dense MoE implements a soft gating mechanism that activates all existing experts during both training and inference. The output can be described as $\sum_{i=1}^{N_{expert}}g_iE_i(x)$, where $E_i(x)$ denotes the output of expert $i$, and $g_i$ is the corresponding gate value. To align with the statistical mixture theory and preserve signal magnitude, MoE usually constrains the sum of gate values to $1.0$ by applying a \emph{Softmax} function for normalization, such that $g=Softmax(x \cdot W_g)$; In contrast, sparse MoE was subsequently proposed by \citet{shazeer2017outrageously} to sparsely activate a small group of experts during each forward propagation, given by $y=\sum_{i\,\in\,TopK(x)} g_iE_i(x)$ where $TopK(x)$ denotes the set of $K$ selected experts with the highest gate values. This strategy achieves better computational efficiency through its sparse activation while still benefiting from a large pool of candidate experts with extensive knowledge. Gshard~\citep{lepikhin2020gshard} was the first to introduce the sparse MoE architecture to Transformer models, after which this powerful and flexible design has begun to be increasingly adopted for training large language models. Llama~\citep{touvron2023llama}, DeepSeek~\citep{liu2024deepseek}, Mixtral~\citep{jiang2024mixtral}, and many other mainstream LLMs are trained based on sparse MoE architectures.

However, despite the success of MoE, some researchers have proposed that the routing mechanisms inside MoE may act as a bottleneck of the whole structure. For example, \citet{zhang2025optimizing} demonstrates that conventional MoE architectures exhibit sudden performance drops during training epochs due to the router's sensitivity to input perturbations; Besides, \citet{ma2024big} emphasizes that conventional gating networks suffer from noise-sensitive and susceptible decision-making that cannot capture nuanced input characteristics. Therefore, some existing studies also focus on boosting the routing ability of MoE. For example, \citet{eo2025mixture} applies a dual-stage routing mechanism consisting of a sentence-level group routing and a token-level expert routing within the selected group; \citet{guo2024dynamic} implements a dynamic mechanism to determine the number of selected experts $k$ instead of treating it as a predefined hyper-parameter; To alleviate the routing imbalance, \citet{zhou2022mixture} enables experts to select top-$k$ tokens instead of enabling tokens to select experts, allowing each expert to maintain a fixed bucket size; \citet{wu2024yuan} designs an attention mechanism for routing, which treats token embeddings as queries and expert embeddings as keys; \citet{harvey2025optimizing} conducts a comparison among six MoE routing mechanisms and concludes that simple routers like a linear module may suffer from overfitting; while complex routers like MLP may suffer from low certainty and thus constrain expert specialization. However, most studies still solely rely on the context of enhancing the scaling routers and their gate value distributions, lacking attention to more complex expert transformations.

\subsection{PEFT and LoRA}

PEFT~\citep{han2024parameter} is a family of techniques designed for adapting large pre-trained language models to downstream tasks while avoiding the prohibitive cost of fully fine-tuning. Instead of updating all billions of parameters, PEFT methods freeze most of the original model weights and only update a small number of trainable parameters or modules, dramatically reducing computational and memory requirements. Popular PEFT methods include Prompt Tuning~\citep{lester2021power}, Prefix Tuning~\citep{li2021prefix}, P-Tuning V2~\citep{liu2022p}, $($IA$)^3$~\citep{liu2022few}, and LoRA~\citep{hu2022lora}, etc. Among these approaches, LoRA has emerged as one of the most influential and widely adopted methods due to its simplicity, effectiveness, and theoretical elegance. Specifically, LoRA approximates the full-rank weight updates $\Delta W_{d \times d}$ through decomposing it into a product of two low-rank trainable matrices $B_{d \times r}$ and $A_{r \times d}$, formulated as $W = W_0 + \Delta W = W_0 + BA$. Here, the rank $r$ is much smaller than the weight matrix dimension $d$. As a result, LoRA significantly reduces the number of trainable parameters.

To enhance LoRA, researchers have explored several technical directions. These include: (1) Improving the initialization of LoRA modules, such as PISSA~\citep{pissa}, MiLoRA~\citep{milora}, and DoRA~\citep{dora}; (2) Optimizing the gradient updates of LoRA, such as LoRA-Pro~\citep{lorapro}, LoRA-GA~\citep{lora-ga}, Riemannian Preconditioned LoRA~\citep{zhang2024riemannian}, and LoRA+~\citep{lora+}; (3) Incorporating regularization and dropout techniques within LoRA, such as RsLoRA~\citep{rslora}, HiddenKey~\citep{hiddenkey}, and LoRA Dropout~\citep{loradropout}; and (4) Integrating multiple LoRA modules into a mixture framework, known as the Mixture of LoRAs or MoE-LoRA, as proposed by \citet{dou2024loramoe,gou2023mixture,mola,luo2024moelora,liu2024adamole}, and \citet{sun2025stronger}. In general, these diverse approaches all aim to address the performance limitation of conventional LoRA through bridging the gap with fully fine-tuning while still maintaining parameter efficiency.

\subsection{MoE-LoRA}
 
Inspired by \citet{adamix}, many researchers have contributed to the field of MoE-LoRA. For example, LoRAMoE~\citep{dou2024loramoe} designs a MoE-LoRA framework for alleviating world knowledge forgetting during fine-tuning LLMs by allocating some experts to specialize only in exploiting pretrained knowledge for task solving; To overcome task conflicts, MoCLE~\citep{gou2023mixture} first groups all instructions and then conducts cluster-level expert routing along with a universal expert; MoLA~\citep{mola} assigns a varying number of experts to different layers and emphasizes that higher layers need more LoRA experts; \citet{luo2024moelora} introduces a contrastive loss into MoE-LoRA to encourage LoRA experts to learn distinct features and improve the efficiency of expert specialization; AdaMoLE~\citep{liu2024adamole} applies a dynamic activation threshold $p$ for each expert and implements a learnable network to adjust it; Instead of training from scratch, MoLE~\citep{mole} discusses the scenario of integrating multiple existing LoRA modules that are already trained on different data sources, and proposes an MoE framework to merge them; \citet{sun2025stronger} proposes an algorithm to integrate Riemannian preconditioners into MoE-LoRA for an enhanced training efficiency and effectiveness; MoRAL~\citep{yang2024moral} focuses on implementing MoE-LoRA for continual lifelong learning; \citet{liu2023moelora} focuses on applying MoE-LoRA in multi-task medical applications; MoLoRA~\citep{shah2026molora} focuses on per-token adapter routing; TalkLoRA~\citep{mu2026talklora} introduces a talking module to low-rank experts for information exchange; CoMoL~\citep{cao2026comol} proposes a core matrix for LoRA, and only implements a routing structure on multiple core matrices instead of multiple low-rank pairs; ReMix~\citep{qiu2026remix} learns routing strategies through reinforcement learning to alleviate the routing weights collapse; SAMoRA~\citep{shi2026samora} uses a semantic-aware routing method in MoE-LoRA for task-adaptive learning; While other works such as \citet{dou2024loramoe,li2024mixlora}, and \citet{sira} focus on load balancing techniques among the low-rank experts. 

All the existing MoE-LoRA works can be roughly separated into two categories: (1) Improving MoE-LoRA by innovative architectural or algorithmic designs; and (2) Applying the MoE-LoRA framework to some specific domains to solve complex multi-task problems. However, none of the works have noticed that the low-rank feature in each LoRA expert can be fully utilized for enhancing MoE-LoRA capabilities through implementing an efficient low-rank transformation for each selected expert beyond the basic gate-value-based scaling in conventional MoE. 

\section{Details of Overall Experimental Setup}
\label{apdx_exp_dtl}

In most of our experiments, we mainly adopt Llama-3.2-3B~\citep{touvron2023llama2} as our foundation model, while for some experiments we also employ Qwen-2.5-3B~\citep{qwen2} and GLM-4-9B~\citep{glm2024chatglm} to demonstrate the generality of our method. We fine-tune all the models on the NVIDIA A100 80GB GPU. To accelerate experiments and reduce resource consumption, we integrate MoE-LoRA trainable modules only into the $q\_proj$, $k\_proj$, $v\_proj$, and $o\_proj$ modules \textbf{only from the first quarter of the model layers}. Specifically, for the 28-layer Llama-3.2-3B, we train the first seven layers (i.e., layer 0-6); For the 36-layer Qwen-2.5-3B, we train the first nine layers (i.e., layer 0-8); While for the 40-layer GLM-4-9B, we train the first ten layers (i.e., layer 0-9). During training, we use a stochastic gradient descent optimizer and set the initial learning rate to 3e-4 with a linear decay scheduler. We fix the system random seed for each run to eliminate the impact of randomization and ensure reproducibility. For initialization, we follow the convention of LoRA to initialize module B to 0. We also initialize our rotation gate $W_\theta$ to 0, and implement a Kaiming uniform initialization for the trainable center vectors $\boldsymbol{q}$. 

To facilitate joint learning, we equally distribute the number of samples across all tasks within every batch. Since the size of the training set varies across different tasks, we choose to \textbf{train each task for the same number of steps rather than epochs}. Consequently, all tasks are treated equally regardless of their difficulty. This setting increases the complexity to conduct joint learning on LLMs, as simple and complicated tasks may exhibit different converging speeds --- when simple tasks show signs of overfitting, complicated tasks may remain unconverged.

\section{Dataset Overview}

\subsection{Mixture of QA Tasks}
\label{qa_intro}

Our mixture of QA tasks includes CommonsenseQA~\citep{talmor2019commonsenseqa}, OpenBookQA~\citep{mihaylov2018can}, and SocialIQA~\citep{sap2019socialiqa}. Specifically, CommonsenseQA consists of more than 12,000 questions requiring multiple types of common sense knowledge; OpenBookQA consists of roughly 6,000 questions from real open book exams on elementary level science facts; while SocialIQA consists of around 38,000 questions about emotional and social reactions during people's everyday life.

\subsection{Mixture of GLUE Tasks}
\label{glue_intro}

The GLUE benchmark~\citep{wang2018glue} targets for evaluating model performances on natural language understanding. It includes nine diverse NLU tasks varying from sentiment classification to grammar checking. To simplify our experiments, here we only select three of the tasks as the mixture: (1) SST-2 which is a sentiment classification task for movie reviews; (2) MRPC which is a paraphrase detection task that determines whether two sentences are semantically equivalent; and (3) QNLI which is a verifying task for candidate QA pairs derived from SQuAD~\citep{rajpurkar2016squad}. 

\subsection{Mixture of MTG Title Generation Tasks}
\label{mtg_intro}

The MTG (Multilingual Text Generation) benchmark~\citep{chen2022mtg} is a human-annotated multilingual text generation benchmark designed for fine-tuning and evaluating language models across diverse languages and tasks. It consists of 400,000 human-annotated samples covering four distinct generation tasks, including the title generation task across five languages: English, Chinese, Spanish, French, and German. Its title generation task, sourced from the ByteCup news corpus, requires models to convert a given article into a condensed, faithful title that preserves the main idea while encouraging readability.

\subsection{Mixture of XNLI Tasks}
\label{xnli_intro}

XNLI~\citep{conneau2018xnli}, known as a cross-lingual version of the Natural Language Inference (NLI) corpus, involves 7,500 human-annotated examples in 15 languages, such as English, Chinese, Spanish, etc. Specifically, every sample in XNLI contains a pair of premise and hypothesis sentences, along with a label indicating if the hypothesis logically follows, contradicts, or is irrelevant to the premise. Notably, to facilitate the experiments and align with MTG's experimental settings, we select the five MTG languages --- English, Chinese, Spanish, French, and German --- from all the 15 languages in XNLI as the language mixture.

\section{Convergence Performance on the Mixture of QA Tasks}
\label{convergence_qa}

\begin{figure}[h]
\centering
\includegraphics[width=\linewidth]{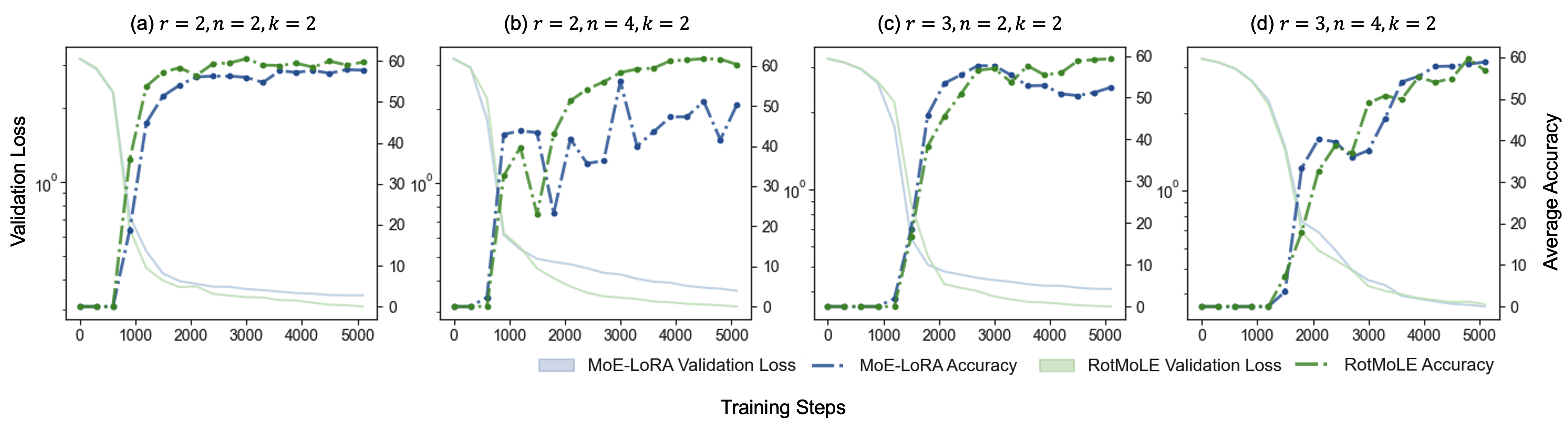}
\caption{Convergence of Jointly Fine-Tuning on Mixed QA Tasks (Qwen-2.5-3B).}
\label{figure2}
\end{figure}

During our fine-tuning experiments on the mixture of QA tasks in Section \ref{qa_section}, we also analyze the converging performances of both the MoE-LoRA baseline and our proposed RotMoLE. Theoretically speaking, RotMoLE implements more trainable parameters and should therefore be more difficult to converge compared to the MoE-LoRA baseline with only a scaling gate. However, experimental results indicate that our RotMoLE does not converge more slowly; while in some cases, it is even faster. Figure \ref{figure2} exhibits this phenomenon. We consider it as an advantage of the rotation gate: using both scaling and rotation to model experts' exploitation is much easier than using scaling alone. Furthermore, RotMoLE also exhibits better converging stability and anti-overfitting ability in experiments under configuration $\{r=2, n=4, k=2\}$ and $\{r=3, n=2, k=2\}$.

\section{Ablation Study}
\label{ablation}

\subsection{Impact of Parameter Size}
\label{ablation_size}

\begin{figure}[ht]
\centering
\includegraphics[width=\linewidth]{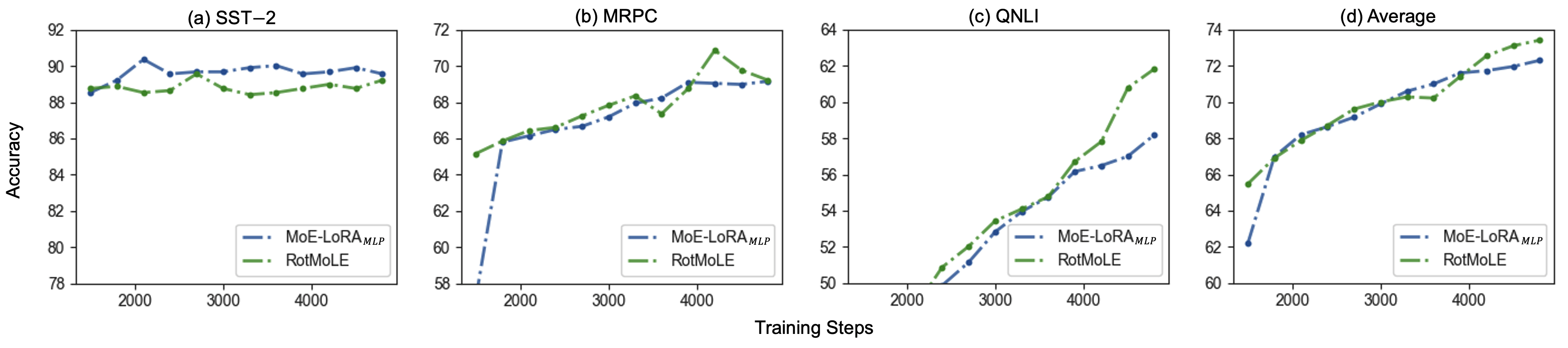}
\caption{Convergence of Jointly Fine-Tuning on Mixed GLUE Tasks by both MoE-LoRA$_{MLP}$ and RotMoLE (Llama-3.2-3B, $r=4, n=5, k=3$).}
\label{figure4}
\end{figure}

Although our method demonstrates a comprehensive improvement over the MoE-LoRA baselines across various scenarios, these gains are built upon the introduction of additional rotation gates $\theta=\Theta(x)$ and center vectors $\boldsymbol{q}$. This raises the question of whether our achieved improvements are due to the increase of trainable parameter counts, rather than our novel rotary architecture design. To clarify this issue, we conduct an experiment on a size-equivalent scaling baseline compared to our proposed RotMoLE. Specifically, we introduce more parameters to the conventional scaling gate, and replace its linear module by a size-equivalent two-layer MLP whose hidden dimension is denoted as $h$. Then we have:
$$2dn+rn=dH+Hn \Rightarrow H=\frac{2dn+rn}{d+n} \approx 2n,$$
considering that $r \ll d$ and $n \ll d$. Consequently, an MLP with a hidden dimension $2n$ serves as the scaling gate of the size-equivalent baseline. We test this size-equivalent baseline under the mixture of three GLUE tasks, which is already elaborated in Section \ref{glue_section}. We use the configuration $\{r=4, n=5, k=3\}$ and present the performance comparisons among the conventional MoE-LoRA baseline, the size-equivalent baseline, and our proposed RotMoLE in Table \ref{mlp_table}. All other settings are the same as in Section \ref{glue_section}. It is observed that the MLP gate performs even worse than the MoE-LoRA baseline with a linear gate, only achieving the best for the sub-task SST-2. We consider it as a phenomenon of overfitting which leads to an unbalanced performance. To illustrate it, we plot the converging traces of MoE-LoRA$_{MLP}$ and RotMoLE in Figure \ref{figure4} from step 1,500. Figure \ref{figure4} indicates that the sub-task SST-2 has already converged, while the other two are still improving. This usually means that SST-2 is much easier than the other two sub-tasks, especially the sub-task QNLI which still has much potential for improvement. MoE-LoRA$_{MLP}$ is overfitted to the easier sub-task SST-2 while lacking sufficient learning of the opportunistic QNLI. This confirms our statement that the MLP gate module is more likely to cause an unbalanced performance and fail to fully utilize the potential of its additional parameters.

\begin{table}[ht]
\centering
\caption{Performance of Jointly Fine-Tuning on Mixed GLUE Tasks by the MoE-LoRA Baseline, the Size-Equivalent MoE-LoRA with MLP Gates, and RotMoLE (Llama-3.2-3B, $r=4, n=5, k=3$).}
\begin{tabular}{lrrrr}
\toprule
\textbf{Method} & \multicolumn{1}{r}{\textbf{SST-2}}    & \multicolumn{1}{r}{\textbf{MRPC}}     & \multicolumn{1}{r}{\textbf{QNLI}}     & \multicolumn{1}{r}{\textbf{Avg.}}     \\ \midrule
MoE-LoRA        & { 88.76}          & { 69.19}          & { 61.63}          & { 73.19}          \\
MoE-LoRA$_{MLP}$        & { \textbf{89.56}} & { 69.16}          & { 58.19}          & { 72.30}          \\
RotMoLE  & { 89.22}          & { \textbf{69.21}} & { \textbf{61.83}} & { \textbf{73.42}} \\ \bottomrule
\end{tabular}
\label{mlp_table}
\end{table}

\subsection{Impact of Model Size}
\label{ablation_base}

We have already examined the effectiveness of our proposal in previous sections by conducting multi-task and multilingual joint learning experiments under Llama-3.2-3B and Qwen-2.5-3B. Now we turn to another foundation model of a different model size, GLM-4-9B. We follow all the experimental settings of Section \ref{qa_section} to conduct comparable experiments of the QA mixture under GLM-4-9B. What is different is that we set the initial learning rate to 9e-4, and train the model for 8,000 steps. The results are shown in Table \ref{glm_table}. An outperformance of our RotMoLE is still observed in both overall and per-task accuracy.

\begin{table}[h]
\centering
\caption{Performance of Jointly Fine-Tuning on Mixed QA Tasks (GLM-4-9B).} 
\begin{tabular}{cclrrrr}
\toprule
\textbf{$r$}        & \textbf{$n, k$}        & \textbf{Method} & \textbf{CSQA}                         & \textbf{OBQA}                         & \textbf{SIQA}                         & \textbf{Avg.}                         \\ \midrule
                    &                        & MoE-LoRA        & { 50.37}          & { 40.20}          & { 45.70}          & { 45.42}          \\
                    & \multirow{-2}{*}{2, 2} & RotMoLE  & \textbf{57.33}                        & \textbf{47.00}                        & \textbf{52.35}                        & \textbf{52.23}                        \\ \cmidrule{2-7} 
                    &                        & MoE-LoRA        & 48.32                                 & 26.80                                 & 42.78                                 & 39.30                                 \\
\multirow{-4}{*}{2} & \multirow{-2}{*}{4, 2} & RotMoLE  & { \textbf{57.82}} & { \textbf{51.80}} & { \textbf{52.71}} & { \textbf{54.11}} \\ \midrule
                    &                        & MoE-LoRA        & { 49.55}          & { 44.00}          & { 49.74}          & { 47.76}          \\
                    & \multirow{-2}{*}{2, 2} & RotMoLE  & \textbf{68.06}                        & \textbf{66.20}                        & \textbf{68.01}                        & \textbf{67.42}                        \\ \cmidrule{2-7} 
                    &                        & MoE-LoRA        & 57.25                                 & 50.60                                 & 55.07                                 & 54.30                                 \\
\multirow{-4}{*}{3} & \multirow{-2}{*}{4, 2} & RotMoLE  & { \textbf{77.48}} & { \textbf{65.60}} & { \textbf{72.16}} & { \textbf{71.75}} \\ \bottomrule
\end{tabular}
\label{glm_table}
\end{table}

\section{Limitations}
\label{limitations}

While our RotMoLE demonstrates significant improvements, we still acknowledge some limitations.  
For example, our evaluations solely focus on the supervised fine-tuning (SFT) scenarios to validate the efficacy of RotMoLE, not incorporating reinforcement learning methods, such as GRPO, in this paper; Secondly, as we point out the phenomenon that various tasks lead to their specific distributions of rotation biases, a further analysis is preferable on whether/how the distribution distinctions are related to the similarities between tasks.



\end{document}